# Probabilistic indirect models for undrained shear strength: addressing significant data missing and variability with advanced imputation and machine learning techniques

Haibin Xiong[1]; Shaoheng Dai[2]; Peng Lan[3]; Xuzhen He[4]; Chenxi Tong[5]; Sheng Zhang[6]; and Daichao Sheng[7]

**Haibin Xiong**, Ph.D. candidate

School of Civil and Environmental Engineering, University of Technology Sydney, Ultimo, NSW 2007, Australia. Email: haibin.xiong@student.uts.edu.au

**Shaoheng Dai**, Postdoctoral

School of Civil and Environmental Engineering, University of Technology Sydney, Ultimo, NSW 2007, Australia. Email: shaoheng.dai@uts.edu.au

**Peng Lan**, Postdoctoral

School of Infrastructure Engineering, Nanchang University, Xuefu Road, 999, Nanchang 330031, China.

**Xuzhen He**, Senior Lecturer **(Corresponding author)**

School of Civil and Environmental Engineering, University of Technology Sydney, Ultimo, NSW 2007, Australia. Email: xuzhen.he@uts.edu.au

**Chenxi Tong**, Associate Professor

School of Civil Engineering, Central South University, Changsha, 410075, China.

Email: cxtong@csu.edu.cn

**Sheng Zhang**, Professor

School of Civil Engineering, Central South University, Changsha, 410075, China.

**Daichao Sheng**, Distinguished Professor

School of Civil and Environmental Engineering, University of Technology Sydney,

Ultimo, NSW 2007, Australia.

**ABSTRACT:** Accurate prediction of undrained shear strength ($s_u$) is crucial for geotechnical design, but is often hampered by substantial uncertainty in traditional empirical methods. This study uses the CLAY/10/7490 global database to develop probabilistic indirect models to predict $s_u$ based on Atterberg limits and piezocone cone penetration (CPTU) measurements. Firstly, the dataset has a high missing data rate and variability. We test three imputation methods – multivariate normal (MN), multiple imputation by chained equations (MICE), and miss forest (MF) – to fill the missing values. To validate their effectiveness, a Probabilistic Extreme Gradient Boosting (PXGB) model is developed, and the imputation methods are evaluated by comparing the PXGB's performance when trained on the imputed datasets against that on the original incomplete data. Secondly, the indirect model is built by integrating a multi-head attention (MHA) mechanism into an artificial neural network (ANN) to enhance information extraction from limited data, which leads to the MHA-based probabilistic neural networks (MHA-PNN) model. The models' performance, alongside a conventional MN-based prediction model, was evaluated using root mean square error (*RMSE*), coefficient of determination ($R^2$), mean absolute percentage error (*MAPE*), conditional interval width ($w_{CI}$), and coverage rate (*CR*). Results demonstrate that the proposed MN-enhanced MHA-PNN model substantially outperforms other models in both prediction accuracy and uncertainty quantification. These findings highlight the potential of this integrated strategy for building robust probabilistic indirect models in geotechnical applications, particularly when confronted with sparse and incomplete datasets.

## 1. Introduction

Undrained shear strength ($s_u$), defined as the maximum shear resistance of soil under conditions where pore water pressure does not dissipate, plays a crucial role in determining the immediate shear capacity of soil. Its accurate assessment remains a significant challenge in geotechnical design, particularly in transient loading scenarios such as seismic events (Belkhatir et al., 2012; Mun et al., 2016), tunnelling construction operations (Soe & Ukritchon, 2023; Ukritchon & Keawsawasvong, 2020), and slope stability assessments (Griffiths & Yu, 2015; Jiang & Huang, 2018; Shogaki & Kumagai, 2008). Current engineering practice primarily employs two approaches for determining $s_u$: (1) laboratory-based measurements such as unconfined compression (UC) and unconsolidated undrained (UU) triaxial testing, and (2) empirical correlations combined with in-situ measurements including field vane shear (FV) and piezocone cone penetration (CPTU) (Yoshimine et al., 1999). Additionally, back-analysis is widely used to estimate unknown soil properties based on laboratory and/or in-situ measurements (Hu et al., 2025b; Zhang et al., 2024).But it requires repeated execution of the forward models, which can be time-consuming and computationally intensive.

All these methods are subject to various sources of geotechnical uncertainty. These uncertainties can be broadly categorized into four types (Hu et al., 2025a; Phoon & Kulhawy, 1999b): (1) inherent variability, resulting from the natural heterogeneity of geological formations; (2) statistical uncertainty, due to the limited data available for estimating soil parameters; (3) measurement error, caused by disturbances during sampling, testing procedures, and other influencing factors (Ding & Loehr, 2019; Lunne et al., 2006; Tanaka, 2000); and (4) transformation uncertainty, introduced when laboratory or in-situ measurements are converted into design parameters using empirical model (Phoon & Kulhawy, 1999a). These compounded uncertainties in $s_u$ are reflected in the highly variable coefficients of variation (COVs) and biases. As reported by Ding and Loehr (2019), COVs range from 0.06-0.56, while biases ($\lambda = s_u / s_{u_ref}$, where $s_{u_ref}$ represents the reference value at the same depth) vary from 0.40 to 1.22 across different testing methods. Such variability fundamentally challenges conventional deterministic approaches in building empirical correlations, underscoring the need for probabilistic frameworks that can effectively quantify uncertainty in $s_u$ predictions.

Recent advances in uncertainty quantification have leveraged Bayesian frameworks with Markov Chain Monte Carlo (MCMC) sampling to update prior distributions using laboratory and field data for geotechnical parameters (Jong & Ong, 2024; Li et al., 2025). However, repeated forward model evaluations in MCMC are computationally expensive. To address this issue, data-driven indirect models have been widely recognized as an effective approach to lighten this computational burden (Lan et al., 2025; Lan et al., 2024; Nguyen et al., 2025; Sun et al., 2023; Wang et al., 2023). Despite their advantages, indirect models are often constructed using relatively small datasets, primarily due to the limited availability of site-specific geotechnical data (Bozorgzadeh et al., 2019; Zhang et al., 2023). To improve the predictive performance, researchers have increasingly turned to multi-source geotechnical datasets to expand the available training data (Beesley & Vardanega, 2020; Yang & Liu, 2024; Zhang et al., 2022; Zhang et al., 2021). For instance, Zhang et al. (2021) utilized the F-CLAY/7/216 and S-CLAY/7/168 datasets (D'Ignazio et al., 2016) to develop extreme gradient boosting and random forest models as indirect models for predicting $s_u$ in soft sensitive clays. Naturally, this has led to growing interest in integrating geotechnical data from around the world for more generalized modelling. One notable effort is by Ching and Phoon (2014b), who compiled the CLAY/10/7490 database with 10 geotechnical features and 7490 rows of data, which are from 251 studies conducted across 30 countries or regions. However, the database bears many missing values and its completion rate is only 34.1%, posing a significant challenge for data-driven modelling. Therefore, effectively utilizing such highly incomplete and highly variable geotechnical big data remains a key and unresolved challenge.

While discarding incomplete records is straightforward, it results in substantial data loss and forfeits valuable information. To address this issue, a range of imputation techniques to assign a value to the missing data have been developed. Basic statistical approaches include mean, median, and mode imputation, whereas more advanced methods encompass linear, polynomial, and spline interpolation (Ye et al., 2024). Interpolation methods, in particular, are widely applied in geotechnical engineering to handle missing data (Kim & Ji, 2022; Xu et al., 2024; Ye et al., 2024). In addition, more sophisticated machine learning (ML) approaches have been explored, such as extreme gradient boosting (XGBoost) (Chian et al., 2023; Xie et al., 2022), k-nearest neighbours (KNN) (Chian et al., 2023; Xie et al., 2022), Bayesian ridge regression (Piao et al., 2025), and miss forest (MF) imputation (Chian et al., 2023), among others. While these methods have shown improved performance in imputing missing data and capturing nonlinear relationships, most of them primarily focus on point estimation. In contrast, statistical imputation methods, which often rely on global data distributions, have the potential to reduce uncertainty more effectively by preserving overall statistical characteristics.

In this paper, we chose: (1) multivariate normal (MN) as a simple imputation baseline method; (2) multiple imputation by chained equations (MICE), which is widely adopted in practice and represents a standard statistical imputation approach; and (3) MF as a representative of advanced ML-based imputation techniques. These methods were compared to provide a comprehensive evaluation across distinct methodological paradigms. To validate their effectiveness, a Probabilistic Extreme Gradient Boosting (PXGB) model is developed, and the imputation methods are evaluated by comparing the PXGB's performance when trained on the imputed datasets against that on the original incomplete data. Additionally, since the imputed data are inferred from limited available information, their quality is highly dependent on the amount and relevance of the observed data. To enhance the model's capability to extract meaningful patterns from such limited inputs, a multi-head attention (MHA) mechanism is incorporated into a conventional artificial neural network (ANN) architecture. Building upon this foundation, we propose a probabilistic modelling framework that integrates imputation strategies with ML techniques. Finally, the proposed framework is applied to the CLAY/10/7490 database and compared with a conventional MN-based prediction model to validate its superiority in terms of both prediction accuracy and uncertainty quantification.

The structure of this paper is as follows: Section 2 introduces the CLAY/10/7490 database, focusing on the prediction of $s_u$ and the imputation of missing data using the MN-based model. Section 3 analyses the missing data patterns within the CLAY/10/7490 database and compares the statistical characteristics of the MN, MICE, and MF methods. In Section 4, the PXGB and MHA-PNN models are developed, and the impact of different imputation techniques on prediction accuracy and uncertainty quantification is assessed. Finally, Section 5 presents the conclusions of this study.

## 2. Prediction and Imputation Based on Multivariate Normal (MN) Model

### *2.1 Missing data distribution in CLAY/10/7490 database*

The CLAY/10/7490 database is a comprehensive clay parameters database and the largest publicly available global database for clay, incorporating site data from 251 studies across 30 countries or regions worldwide (Ching & Phoon, 2014b). However, its completeness rate is around 34.1%. To improve the database's utility, it is essential to first analyse the distribution of the missing data. Figure 1 provides a visualization of data completeness across 8 parameters: (1) Atterberg limit parameters: liquid limit (*LL*), plasticity index (*PI*), liquidity index (*LI*); (2) CPTU parameters: normalized cone tip resistance ($(q_t-\sigma_v)/\sigma'_v$), effective cone tip resistance ($(q_t-u_2)/\sigma'_v$), normalized excess pore pressure ($(u_2-u_0)/\sigma'_v$), and pore pressure ratio $B_q$; and (3) normalized mobile undrained shear strength ($s_u(\text{mob})/\sigma'_v$), which represents the mobilized in situ shear strength under undrained failure, converted from $s_u$ obtained from various tests using transformation models (Ching & Phoon, 2014a, 2014b). The figure illustrates the distribution of missing data, with white areas representing missing values and black areas indicating available data. The extent of missing data varies across different parameters, with CPTU parameters exhibiting particularly high levels of incompleteness. Other parameters in the database, such as preconsolidation stress ($\sigma_p$) and sensitivity ($s_t$), are often difficult to obtain in practice. Therefore, to develop a probabilistic indirect model using easily accessible variables for predicting $s_u(\text{mob})/\sigma'_v$, we deliberately focused on these seven input features (*LL*, *PI*, *LI*, and the four CPTU parameters). Notably, while this global database offers valuable insight into general trends and uncertainty across a wide range of clays, it is intended as a basis for first-order estimates rather than a substitute for site-specific predictions.

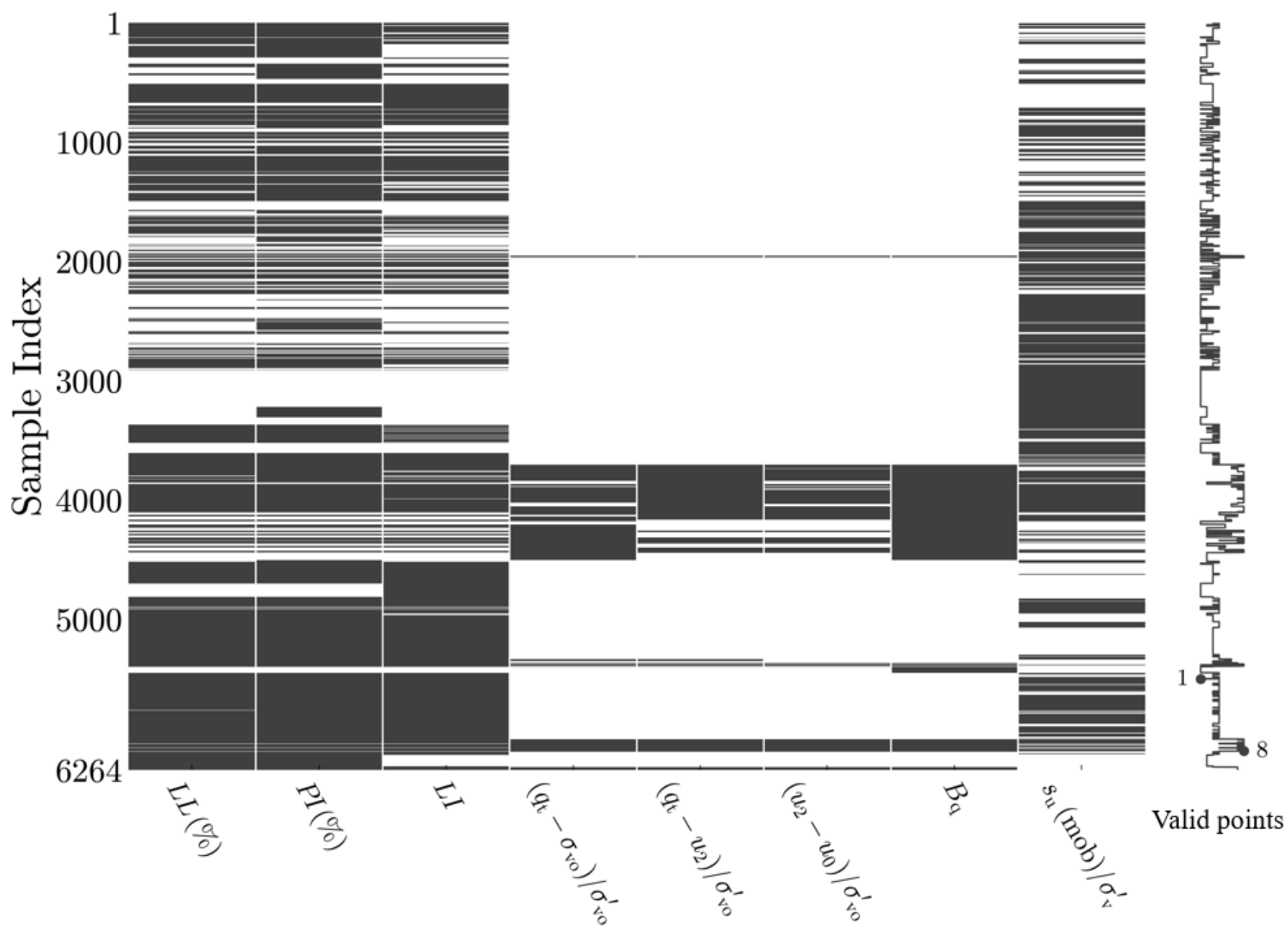

**Fig. 1.** Missing data distribution across 8 parameters for the CLAY/10/7490 database

Table 1 provides a more detailed statistical summary, revealing that $s_u(mob)/\sigma'_v$ and Atterberg limit parameters have relatively lower missing rates (approximately 40% to 50%), whereas CPTU parameters suffer from severe data missing, with some variables missing over 90% of their values.

**Table 1.** The statistics of the CLAY/10/7490 database

| Type | Variable | Mean | COV | Min | Max | Points | Missing (%) |
|---|---|---|---|---|---|---|---|
| Atterberg Limit | *LL* | 68.649 | 0.811 | 18.067 | 550.000 | 4057 | 45.8 |
| | *PI* | 40.210 | 1.095 | 1.895 | 493.000 | 4503 | 39.9 |
| | *LI* | 1.005 | 0.795 | -0.775 | 6.455 | 3795 | 49.3 |
| CPTU | $(q_t\text{-}\sigma_v)/\sigma'_v$ | 9.450 | 1.233 | -1.041 | 110.561 | 880 | 88.3 |
| | $(q_t\text{-}u_2)/\sigma'_v$ | 5.655 | 1.460 | 0.357 | 108.201 | 752 | 90.1 |
| | $(u_2\text{-}u_0)/\sigma'_v$ | 4.732 | 1.342 | -0.065 | 117.327 | 668 | 91.1 |
| | $B_q$ | 0.569 | 0.358 | 0.000 | 1.172 | 1017 | 86.4 |
| $s_u$ | $s_u(mob)/\sigma'_v$ | 0.541 | 1.670 | 0.019 | 27.755 | 3779 | 49.5 |

Beyond data completeness, the table reveals significant variability across parameters, with CPTU parameters showing particularly high variation, as $(q_t – \sigma_v)/\sigma'$, $(q_t – u_2)/\sigma'_v$, and $(u_2\text{-}u_0)/\sigma'_v$ exceeding a COV of 1.2. The highest variability is observed in $s_u(\text{mob})/\sigma'_v$ (COV = 1.670), reflecting substantial variability in these parameters. Such variations pose challenges for analysis and modelling. To enhance the utility of the database, implementing strategic data imputation and adopting robust approaches to handle variable uncertainty are crucial for ensuring reliable geotechnical applications.

### *2.2 MN-based prediction model for $s_u(mob)/\sigma'_v$*

Given the high missingness and variability observed in the database, an effective probabilistic framework is essential for estimating missing values while preserving the underlying statistical relationships among parameters. The MN-based model offers a structured approach to achieving this; however, as all parameters exhibit non-Gaussian distribution, as shown in Figure 2, a transformation to approximate normality is necessary before constructing the MN-based model. The Johnson system of distributions, comprising three families (log-normal (SL), unbounded (SU), and bounded (SB)) provides a flexible method for this transformation (Slifker & Shapiro, 1980).

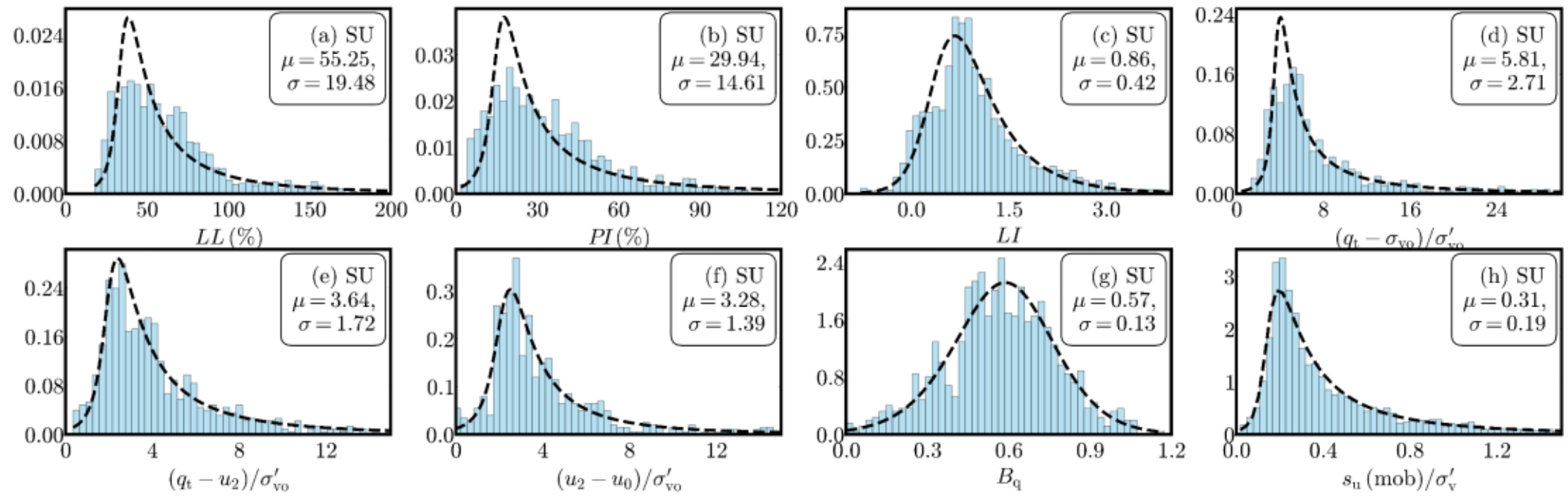

**Fig. 2.** Distributions of all parameters and their Johnson fitting

The mathematical formulations are as follows:

$$Johnson-SU\text{: } f(x)=\frac{a_x}{a_y}\cdot\frac{\phi\left(b_x+a_x\cdot\sin^{-1}\left(\left(x-b_y\right)/a_y\right)\right)}{\sqrt{1+\left(1+\left(x-b_y\right)/a_y\right)^2}} \tag{1}$$

$$Johnson-SB\text{: } f(x)=\frac{a_x}{a_y}\cdot\frac{\phi\left(b_x+a_x\left[\ln\left(\left(x-b_y\right)/a_y\right)-\ln\left(1-\left(x-b_y\right)/a_y\right)\right]\right)}{\left(\left(x-b_y\right)/a_y\right)\left(1-\left(x-b_y\right)/a_y\right)} \tag{2}$$

$$Johnson-SL\text{: } f(x)=\frac{a_x}{a_y}\cdot\frac{\phi\left(b_x+a_x\ln(a_y)+a_x\ln\left(\left(x-b_y\right)/a_y\right)\right)}{\left(x-b_y\right)/a_y} \tag{3}$$

where $a_x$ , $b_x$ , $a_y$ , and $b_y$ are the parameters of Johnson distributions, $x$ is the real value of each variable, and $\phi(\cdot)$ is the value of the probability density function (PDF) for the standard normal distribution.

Following the approach outlined by Slifker and Shapiro (1980), we classify all parameters under the SU family and estimate the eight transformation model parameters, as summarized in Table 2. Figure 2 illustrates the fitted results alongside the histograms of all parameters, including the corresponding means ($\mu$) and standard deviations ($\sigma$). Despite the varying degrees of skewness and kurtosis, the fitted curves align well with the true distributions, verifying that the Johnson-SU distribution is appropriate for transforming the parameters into normal space.

**Table 2.** Fitted parameters of the Johnson distributions for CLAY/10/7490 database

| Variable | $a_x$ | $b_x$ | $a_y$ | $b_y$ | Family |
|---|---|---|---|---|---|
| $LL$ | 0.75 | -1.18 | 7.29 | 33.99 | SU |
| $PI$ | 0.70 | -1.10 | 4.88 | 15.05 | SU |
| $LI$ | 1.50 | -1.19 | 0.66 | 0.28 | SU |
| $(q_t-\sigma_v)/\sigma'_v$ | 0.64 | -1.10 | 0.71 | 3.68 | SU |
| $(q_t-u_2)/\sigma'_v$ | 0.77 | -1.14 | 0.71 | 1.96 | SU |
| $(u_2-u_0)/\sigma'_v$ | 0.78 | -0.87 | 0.82 | 2.15 | SU |
| $B_q$ | 2.52 | 0.26 | 0.47 | 0.63 | SU |
| $s_u(\text{mob})/\sigma'_v$ | 0.80 | -1.44 | 0.06 | 0.14 | SU |

Next, to develop the MN-based prediction model, we adopt the framework for constructing the distribution as outlined in previous studies (Ching et al., 2014; He et al., 2025). This approach involves transforming the original parameters into a normal space, followed by estimating the statistical correlations among the transformed parameters to predict the probability distributions of the target variables.

The first step involves selecting appropriate transformation functions to map the original parameter into a normal space. Based on the distribution patterns observed in Figure 2 and the statistical results in Table 1, we make the following determinations: Since $s_u(\text{mob})/\sigma'_v$ is always positive and predominantly concentrated at lower values, a logarithmic transformation is the most suitable choice. Similarly, as $B_q$ falls within the range (0, 1.2), an inverse sigmoid function is appropriate. For all other parameters, the inverse hyperbolic function is applied. Accordingly, the transformation functions are defined as follows:

$$T_i = \begin{cases} \sinh^{-1}(X_i), & \text{for } i = 1, 2, \ldots, 6 \\ \text{sigmoid}^{-1}(X_7 / 1.2), & \text{for } i = 7 \\ \ln(X_8), & \text{for } i = 8 \end{cases} \tag{4}$$

where $T_1$ to $T_8$ represent the transformed parameters, with $X_1 = LL$, $X_2 = PI$, $X_3 = LI$, $X_4 = (q_t\text{-}\sigma_v)/\sigma'_v$, $X_5 = (q_t\text{-}u_2)/\sigma'_v$, $X_6 = (u_2\text{-}u_0)/\sigma'_v$, $X_7 = B_q$, and $X_8 = s_u(\text{mob})/\sigma'_v$. To further evaluate the effectiveness of the transformation, Figure 3 presents the distributions of the transformed parameters ($T_1$ to $T_8$) and their correlations. The diagonal figures demonstrate that $T_1$ to $T_8$ exhibit normality. Furthermore, normality tests, including the quantile-quantile (Q-Q) plots and Anderson-Darling tests, confirm that the transformed parameters follow a marginal normal distribution, validating the suitability of the transformations. The estimated mean ($\mu_i$) and standard deviations ($\sigma_i$) of these transformed variables are listed in Column 2-3 of Table 3. The Pearson correlation coefficients $\rho_{jk}$ between the $T_j$ and $T_k$ is estimated using the bootstrapping method suggested by Ching et al. (2014) and are listed in Table 3. The last column of Figure 3 shows that the Atterberg limit parameters exhibit little to no correlation with $s_u(\text{mob})/\sigma'_v$, whereas the CPTU parameters have a strong linear relationship with it.

**Table 3.** Parameters of the MN-based prediction model for $s_u(\text{mob})/\sigma'_v$ prediction

| $T_i$ | $\mu_i$ | $\sigma_i$ | $\rho_{ij}$ | | | | | | | |
|---|---|---|---|---|---|---|---|---|---|---|
| | | | $T_1$ | $T_2$ | $T_3$ | $T_4$ | $T_5$ | $T_6$ | $T_7$ | $T_8$ |
| $T_1$ | 4.71 | 0.35 | 1.00 | 0.96 | -0.23 | 0.00 | 0.02 | 0.10 | 0.16 | 0.07 |
| $T_2$ | 4.09 | 0.48 | 0.96 | 1.00 | -0.33 | -0.07 | -0.04 | 0.05 | 0.18 | 0.03 |
| $T_3$ | 0.78 | 0.31 | -0.23 | -0.33 | 1.00 | -0.11 | -0.17 | -0.14 | 0.01 | -0.13 |
| $T_4$ | 2.46 | 0.41 | 0.00 | -0.07 | -0.11 | 1.00 | 0.86 | 0.77 | -0.39 | 0.80 |
| $T_5$ | 2.00 | 0.43 | 0.02 | -0.04 | -0.17 | 0.86 | 1.00 | 0.49 | -0.61 | 0.69 |
| $T_6$ | 1.90 | 0.38 | 0.10 | 0.05 | -0.14 | 0.77 | 0.49 | 1.00 | 0.17 | 0.65 |
| $T_7$ | -0.10 | 0.45 | 0.16 | 0.18 | 0.01 | -0.39 | -0.61 | 0.17 | 1.00 | -0.27 |
| $T_8$ | -1.17 | 0.52 | 0.07 | 0.03 | -0.13 | 0.80 | 0.69 | 0.65 | -0.27 | 1.00 |

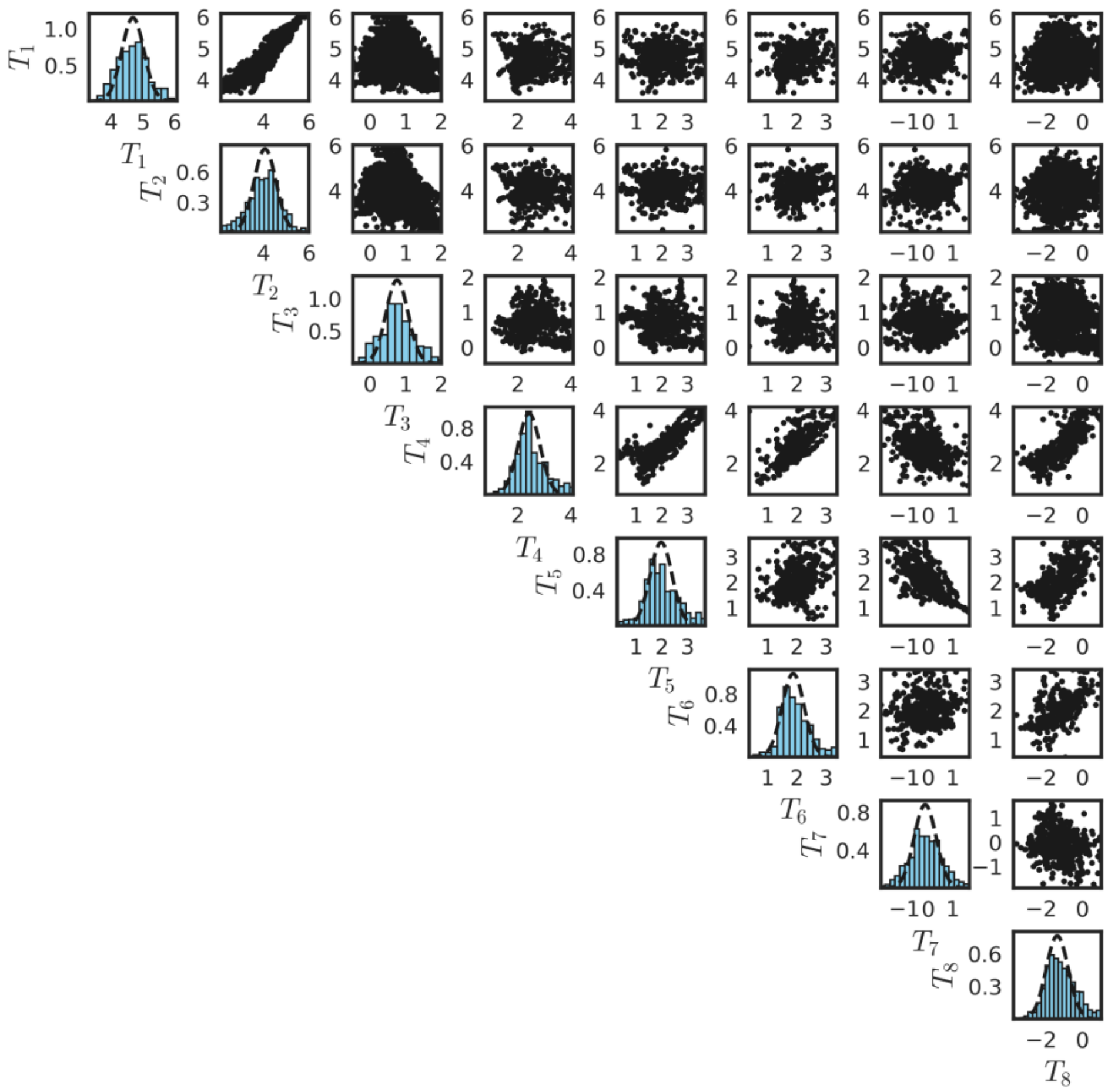

**Fig. 3.** Distributions and correlations of transformed parameters ($T_1$-$T_8$)

All the transformed parameters, $T_1$ to $T_8$, follow a multivariate normal distribution, then the conditional distribution $p(T_i^{\text{pre}}|T_j, T_k, ...)$ is also a normal distribution and can serve as a predictive model. Here, $T_i^{\text{pre}}$ represents the predicted distribution for the transformed variable $i$. When $i$ = 8, it corresponds to the predicted distribution of the transformed $s_\text{u}$(mob)/$\sigma'_\text{v}$. $T_j$, $T_k$, ... are known parameters. For example, for a soil sample, three values, $X_2 = PI$, $X_5 = (q_\text{t}\text{-}u_2)/\sigma'_\text{v}$, and $X_6 = (u_2\text{-}u_0)/\sigma'_\text{v}$ are available. Then, $T_2$, $T_5$, and $T_6$ can be integrated into the conditional distribution to have a predictive distribution for $T_8^{\text{pre}}$, i.e., prediction regarding $s_\text{u}$(mob)/$\sigma'_\text{v}$. The conditional distribution can be explicitly expressed as

$$T_i^{\text{pre}} \sim \text{N}\left(\mu_i^{\text{pre}}, \left(\sigma_i^{\text{pre}}\right)^2\right) \tag{5}$$

$$\mu_i^{\text{pre}} = \mu_i + \begin{bmatrix} c_{ij} & c_{ik} & \cdots \end{bmatrix} \begin{bmatrix} c_{jj} & c_{jk} & \cdots \\ c_{jk} & c_{kk} & \\ \vdots & & \ddots \end{bmatrix}^{-1} \begin{bmatrix} T_j - \mu_j \\ T_k - \mu_k \\ \vdots \end{bmatrix} \tag{6}$$

$$\left(\sigma_i^{\text{pre}}\right)^2 = c_{ii} - \begin{bmatrix} c_{ij} & c_{ik} & \cdots \end{bmatrix} \begin{bmatrix} c_{jj} & c_{jk} & \cdots \\ c_{jk} & c_{kk} & \\ \vdots & & \ddots \end{bmatrix}^{-1} \begin{bmatrix} c_{ij} \\ c_{ik} \\ \vdots \end{bmatrix} \tag{7}$$

$$\boldsymbol{C} = \begin{bmatrix} c_{11} & c_{12} & \cdots \\ c_{21} & c_{22} & \\ \vdots & & \ddots \end{bmatrix} = \begin{bmatrix} \sigma_1 & 0 & \cdots \\ 0 & \sigma_2 & \\ \vdots & & \ddots \end{bmatrix} \begin{bmatrix} 1 & \rho_{12} & \cdots \\ \rho_{21} & 1 & \\ \vdots & & \ddots \end{bmatrix} \begin{bmatrix} \sigma_1 & 0 & \cdots \\ 0 & \sigma_2 & \\ \vdots & & \ddots \end{bmatrix} \tag{8}$$

where $\boldsymbol{C}$ denotes the covariance matrix, which can be calculated from the standard deviations ($\sigma_i$) and the Pearson correlation coefficients $\rho_{jk}$ as in Eq.(8). The estimated mean ($\mu_i$) and standard deviations ($\sigma_i$) are those in Table 3. $\mu_i^{\text{pre}}$ and $\sigma_i^{\text{pre}}$ are just calculated intermediate values. This predictive model is then used for $s_\text{u}$(mob)/$\sigma'_\text{v}$ on the CLAY/10/7490 dataset, and the results are depicted in Figure 4. Because for different soil samples, we have different numbers of available ($T_j$, $T_k$,...) as shown in Figure 1. The results are plotted on different figures depending on how many inputs are available. The vertical lines indicate the 95% confidence intervals, and scatter plots illustrate the comparison between predicted mean values and corresponding actual values on a logarithmic scale.

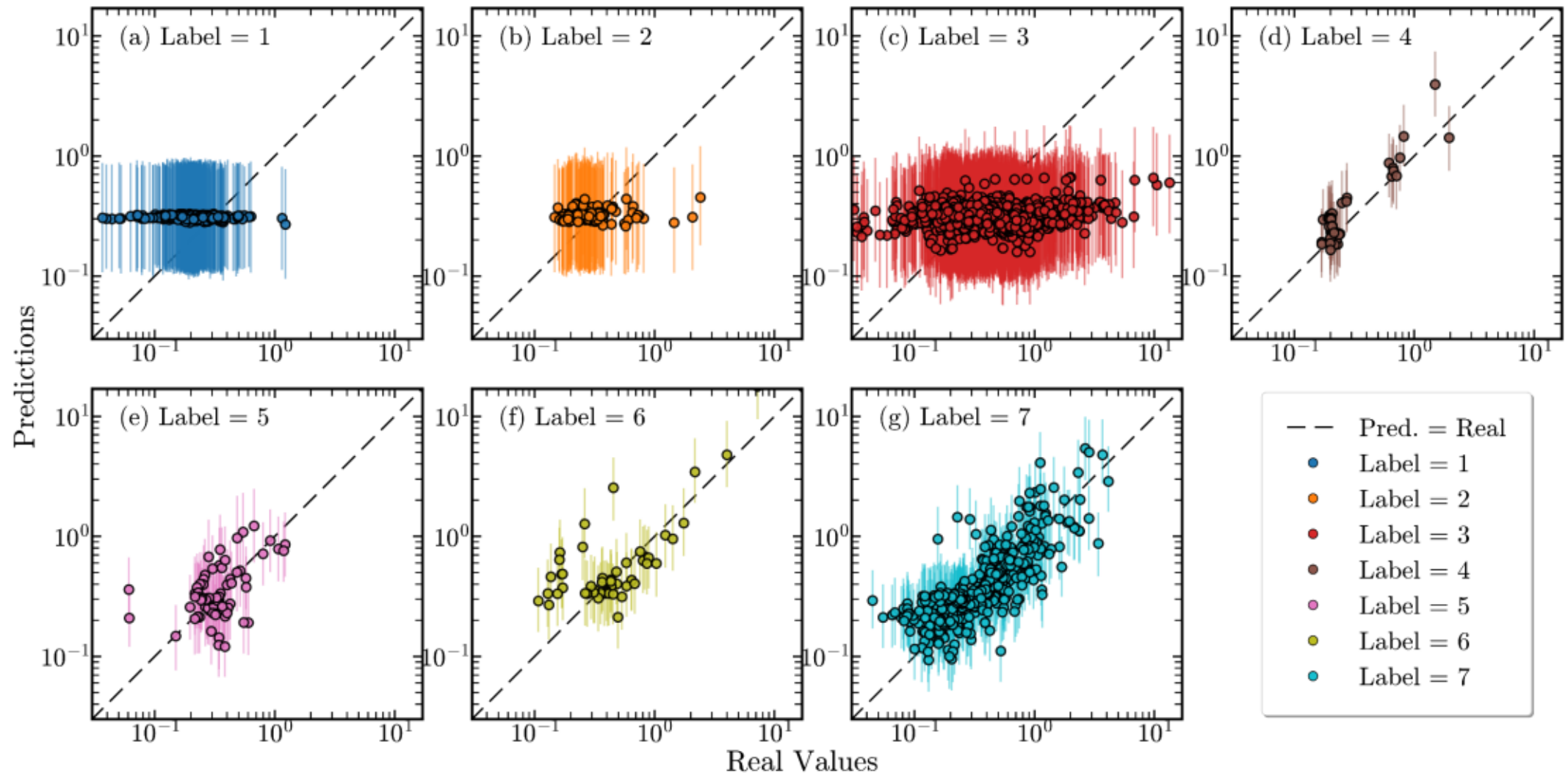

**Fig. 4.** Prediction for $s_u(\mathrm{mob})/\sigma'_v$ based on the MN-based prediction model (Labels 1-7 represent the number of available inputs, ranging from 1 to 7)

Notably, in Figure 4(a), when only a single input feature is available, the model tends to generate nearly identical predictions, which is inaccurate. This suggests that with an insufficient number of input features, the model lacks the necessary information to distinguish between varying cases, resulting in oversimplified and inaccurate predictions. Figures 4(b)-(c) illustrate the prediction results when the number of input features increases to two and three. Their predicted mean values mostly fall between 0.1 and 0.8, while the real values cover a wider range, which still has low accuracy. The main reason for this issue is that when the number of available input features does not exceed three, the database mainly consists of Atterberg limit parameters, with very few CPTU parameters. Consequently, when the model predominantly relies on the Atterberg limit parameters as the main input feature, it fails to capture the true probabilistic relationship of $s_u(\text{mob})/\sigma'_v$, leading to inaccurate predictions.

Figures 4(d)-(g) illustrate the prediction results of the MN-based prediction model using four to seven input features. Compared to Figures 4(a)-(c), the predictions demonstrate a stronger alignment with the diagonal reference line. Additionally, the confidence intervals are generally narrower, indicating reduced uncertainty and greater reliability in the model's predictions. This improvement is due to incorporating CPTU parameters, highlighting the importance of selecting informative features. Without them, the model struggles to accurately capture the true underlying distribution of the target variable. Note that the number of data points in Figures 4(a)-(g) varies due to differences in sample sizes associated with each label (available input features).

Overall, the MN-based prediction model provides a statistically rigorous approach for predicting $s_u(\text{mob})/\sigma'_v$, reducing dependence on less reliable empirical methods, especially in the presence of missing data. The results underscore the importance of incorporating relevant input features to enhance predictive accuracy. As more informative variables are included, the model better captures probabilistic relationships, leading to narrower confidence intervals and improved agreement with observations. Additionally, its ability to explicitly account for uncertainty makes it well-suited for geotechnical applications where data sparsity is a challenge.

***2.3 MN-based imputation method***

As described in Section 2.2, the MN-based prediction model utilizes the linear statistical relationships among transformed variables to construct their multivariate normal distribution, which is then applied to predict the probabilistic distributions of the target variables. While the results in Figures 4(d)-(g) demonstrate strong predictive performance, this section aims to assess whether imputing missing input parameters leads to improved results compared to using the original incomplete database. It also serves as a foundation for developing new models that require a complete database for training. Notably, only input data will be imputed, while target data will remain unchanged to avoid potential issues in subsequent prediction and validation.

In the previous section, we used the conditional distribution as a predictive model (MN-based prediction model) for $s_u(\text{mob})/\sigma'_v$ only, i.e., $T_i^{\text{pre}}$ with $i$ = 8. If we choose $i$ as the indices of missing parameters, we can also have predictions for the missing parameters, referred to as the MN-based imputation method, specifically:

$$\boldsymbol{T}_{\text{miss}}^{\text{pre}} \sim \text{N}\left(\boldsymbol{\mu}_{\text{miss}}^{\text{pre}}, \left(\boldsymbol{\sigma}_{\text{miss}}^{\text{pre}}\right)^2\right) \tag{9}$$

$$\boldsymbol{\mu}_{\text{miss}}^{\text{pre}} = \boldsymbol{\mu}_{\text{miss}} + \boldsymbol{C}_{\text{miss,non-miss}} \boldsymbol{C}_{\text{non-miss,non-miss}}^{-1} \left(\boldsymbol{T}_{\text{non-miss}} - \boldsymbol{\mu}_{\text{non-miss}}\right) \tag{10}$$

$$\boldsymbol{\sigma}_{\text{miss}}^{\text{pre}} = \boldsymbol{\sigma}_{\text{miss}} + \boldsymbol{C}_{\text{miss,non-miss}} \boldsymbol{C}_{\text{non-miss,non-miss}}^{-1} \boldsymbol{C}_{\text{non-miss,miss}} \tag{11}$$

where the subscript miss denotes the indices of missing values, and the subscript non-miss represents the indices of non-missing (observed) values. Consequently, for a given sample, if a non-missing value is present in the input, all missing values for that sample can be inferred. This prediction is a distribution; we choose the mean as the imputed value. After imputation, the imputed dataset is shown in Figure 5 and compared with the original dataset. Notably, the imputation method effectively preserves the statistical relationships among all variables. This aligns with the fundamental characteristics of the multivariate normal model, providing an initial validation of the effectiveness of the imputation process. The figure presents the pairwise relationships among $T_1$-$T_7$, where the upper triangle displays scatter plots of the original database, the lower triangle shows scatter plots of the database obtained using imputation method based on multivariate normal model (hereafter referred to as the MN database), and the diagonal elements display histograms of the MN database. Comparatively, the variables in the MN database retain the original distribution characteristics while slightly expanding the range of variable distributions, all while maintaining a high degree of consistency with the original database. Furthermore, the MN database accurately reproduces the relationships among variables, thereby reinforcing our confidence in utilizing this database for further analyses.

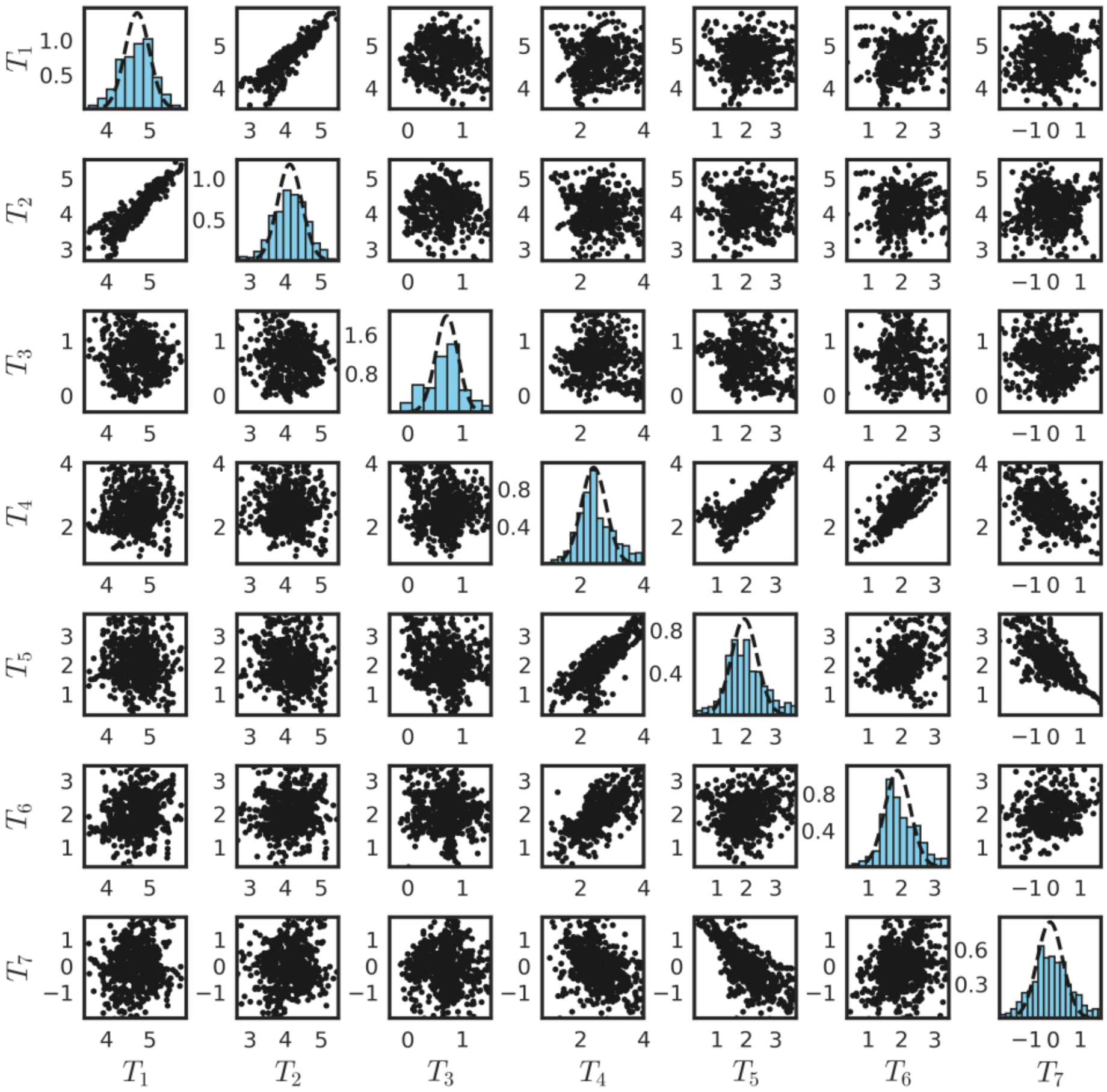

**Fig. 5.** Distribution of transformed variables in the original (upper triangular) and imputed (lower triangular) database

After comparing their distribution characteristics, the next focus is on whether the MN database can be effectively used to build a similar multivariate model and enhance the predictive performance. Based on the results in Figure 4, which show prediction differences with varying available inputs, we categorized the MN database by the original number of available input features and compared its performance with the original database. Figure 6 illustrates this, with the original database shown by blue squares and the new model by red circles. As the number of input features increases, the predictive performance of both models gradually improves. This observation aligns with the trend observed in Figure 4, where an increase in the number of input features led to improved model performance.

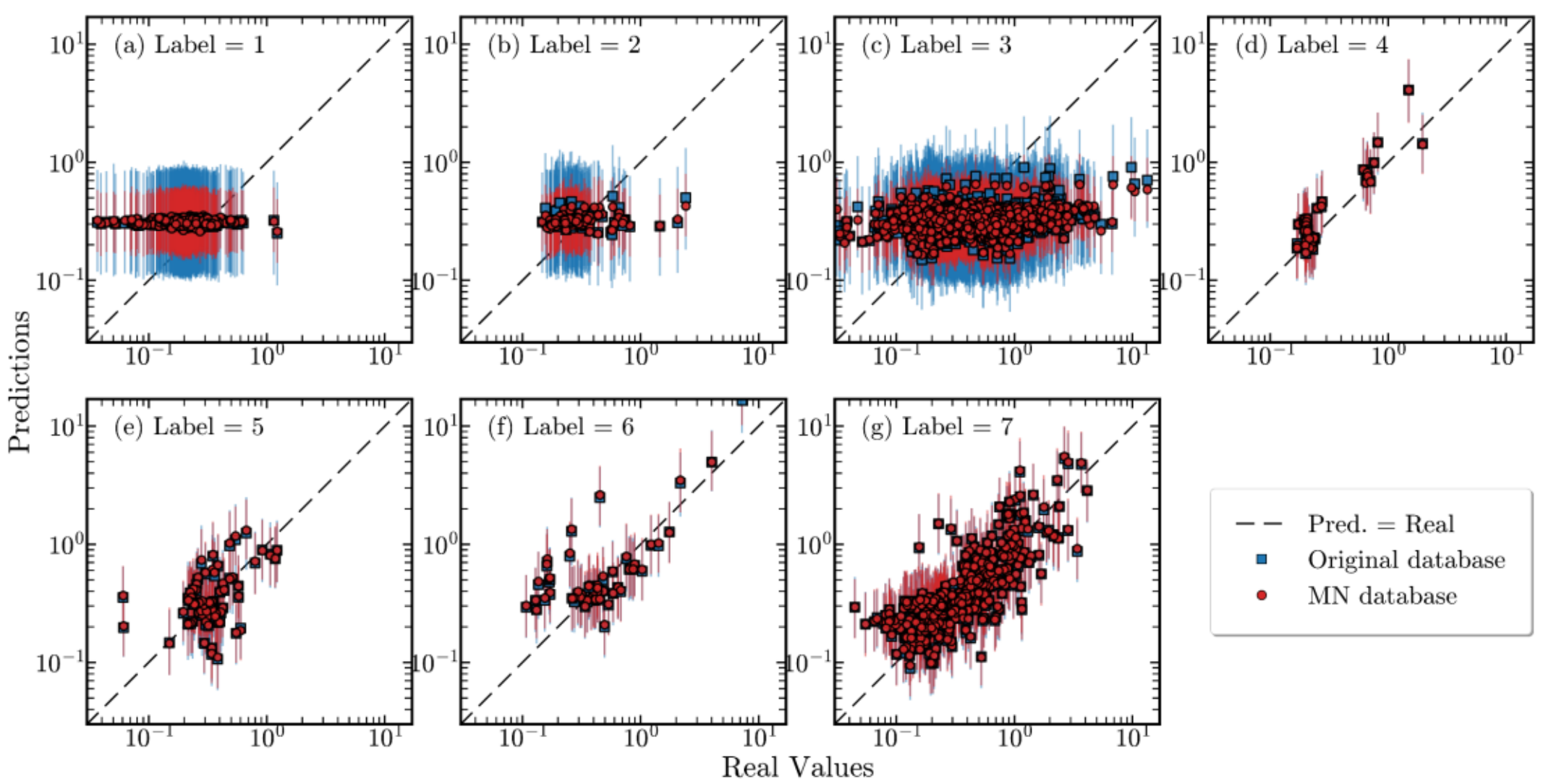

**Fig. 6.** Comparison of prediction results using the original and MN database (Labels 1-7 represent the number of available inputs, ranging from 1 to 7)

Compared to the original results, the predictions obtained using the MN database in Figures 6(a)-(c) exhibit a more concentrated distribution. However, the results presented in Figures 6(a)-(g) remain largely consistent overall. This suggests that imputing the original data with the MN-based imputation method does not lead to a significant improvement in prediction accuracy. The primary reason for this is that both the MN-based imputation method and the MN-based prediction model are based on multivariate normal distribution theory. Therefore, the MN-based imputation method does not provide additional independent information to improve the model's predictive capability. Consequently, the model produces similar outcomes, which is expected due to the overlap in the theoretical foundations of both approaches. Hence, the subsequent chapters will explore alternative methods and models aimed at further improving prediction accuracy.

## 3. Imputation Based on Multiple Imputation and Miss Forest

### *3.1 Missing data patterns for CLAY/10/7490 Database*

As described in Section 2.3, we have successfully imputed missing data in the Clay/10/7490 database using the MN-based imputation method, while preserving the statistical relationships between different variables. However, the predictive performance has not improved. Therefore, in this chapter, we will explore alternative data imputation methods, including various statistical and machine learning techniques. Given the high missing data rates characteristic of the Clay/10/7490 database, it is crucial to first analyse the underlying missing data mechanisms before investigating alternative approaches. Common missing data mechanisms include missing completely at random (MCAR), missing at random (MAR), and missing not at random (MNAR). The distinction between these mechanisms lies in the relationship between the missing and observed (or unobserved) data: MCAR is entirely random, MAR depends on observed data, and MNAR is influenced by unobserved data.

Figure 7 illustrates the correlation matrix of missing data patterns among the transformed variables ($T_1$-$T_8$), visualized as a heatmap. The colour scale ranges from deep red (indicating strong positive correlation) to deep blue (indicating strong negative correlation), representing the strength of relationships between the missing data patterns of different variables. It is observed that the transformed Atterberg limit parameters ($T_1$, $T_2$, and $T_3$) exhibit a strong positive correlation (> 0.6), suggesting that their missing data patterns are likely influenced by similar factors, such as sampling limitations or measurement inadequacies. Since these factors are easily observable in field, their missing data mechanism is best characterized as MAR. Similarly, the transformed CPTU parameters ($T_4$, $T_5$, $T_6$, and $T_7$) display stronger correlations (> 0.8) and, as presented in Table 3, exhibit a high missing rate (> 86%). This suggests that their missing values are primarily influenced by common factors, such as equipment limitations, further supporting the assumption that their missing data mechanism follows the MAR.

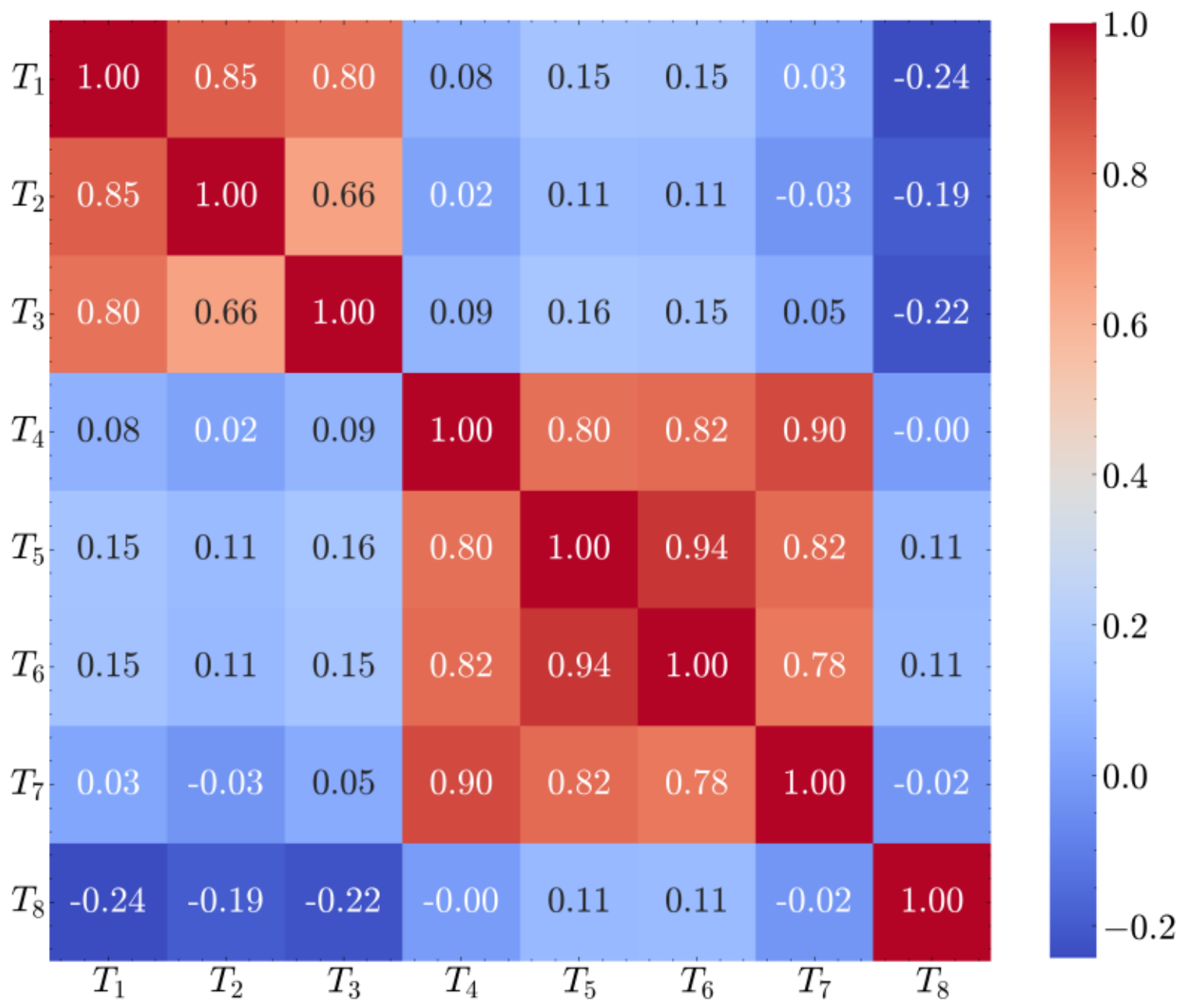

**Fig. 7.** Correlation matrix of missing data patterns among transformed variables

In contrast, $T_8$ shows weaker correlations with other variables, and in some cases, even negative correlations (e.g., -0.2 with $T_1$, $T_2$, and $T_3$). This suggests that the data collection method or source of $T_8$ is more independent, and its missing data mechanism may be closer to MNAR or have more complex characteristics. However, as $T_8$ serves as the target variable, imputing its missing values could introduce bias. Therefore, consistent with the previous method, we choose not to impute $T_8$. As a result, its more complex missing data mechanism does not influence the overall validity of our findings.

Given that the missing data mechanism for these variables ($T_1$-$T_7$) aligns with the MAR assumption, we employ both statistical and machine learning-based imputation techniques to improve data completeness. Specifically, we will apply multiple imputation by chained equations (MICE) and miss forest (MF), as both methods effectively utilize inter-variable correlations to enhance imputation accuracy. By these methods, we aim to mitigate the impact of missing data while preserving the underlying statistical structure of the database.

### *3.2 Multiple Imputation by Chained Equations (MICE)*

MICE is a robust statistical technique for addressing multivariate missing data, widely used due to its flexibility and effectiveness in imputing missing values. This section introduces the MICE method and provides a brief comparison with the MN-based imputation method outlined in Section 2.3. The MN-based imputation method relies on the assumption that all data follow a multivariate normal distribution, using this theoretical framework to estimate and impute missing data. In contrast, MICE offers greater flexibility by modelling each variable's missing values based on its own distribution, without assuming a multivariate normal distribution. Moreover, MICE do not require variables to be transformed into an approximately normal distribution before imputation, making it more adaptable for handling diverse data types.

The MICE process is iterative, using regression models to impute missing data sequentially. In each iteration, other observed variables serve as predictors to estimate the missing values of the current variable. This approach models missing data based on the specific distribution of each variable, providing a more accurate imputation method. The main steps of the MICE procedure for imputing missing data are as follows: (1) Use a simple imputation method (e.g., mean imputation) to fill in the missing values as initial guesses. (2) Select a variable with missing values as the target variable and use other variables as predictors. Apply a regression model (such as linear regression, logistic regression, or other more complex models) to predict and impute the missing values of the target variable. (3) Repeat step (2) for all variables with missing values, using newly imputed values to refine subsequent imputations. (4) Iterate steps (2) and (3) multiple times until the imputed values converge, ensuring stability in the results. (5) Repeat this process several times (typically 5 to 10 iterations), generating multiple imputed databases. These databases are then analysed independently, and the results are combined to reduce uncertainty in the imputation process.

Initially, based on the results from the MN-based imputation method, each sample in the database was labelled according to the number of available input features, with labels ranging from 0 to 7. The original database, containing 4,973 samples, excluded data points with a label of 0. For comparative analysis, three subsets were created based on label thresholds: The first subset (Subset 1), with 4,316 samples, had labels greater than 1; the second subset (Subset 2), with 3,747 samples, contained labels greater than 2; and the third subset (Subset 3), with 809 samples, consisted of labels greater than 3.

The MICE method was applied for 10 iterations to each of these subsets. The results are presented in Figure 8 as histograms. The top row of Figure 8 corresponds to the original dataset (excluding data points labelled as 0), while the second to fourth rows correspond to Subsets 1 to 3, respectively.

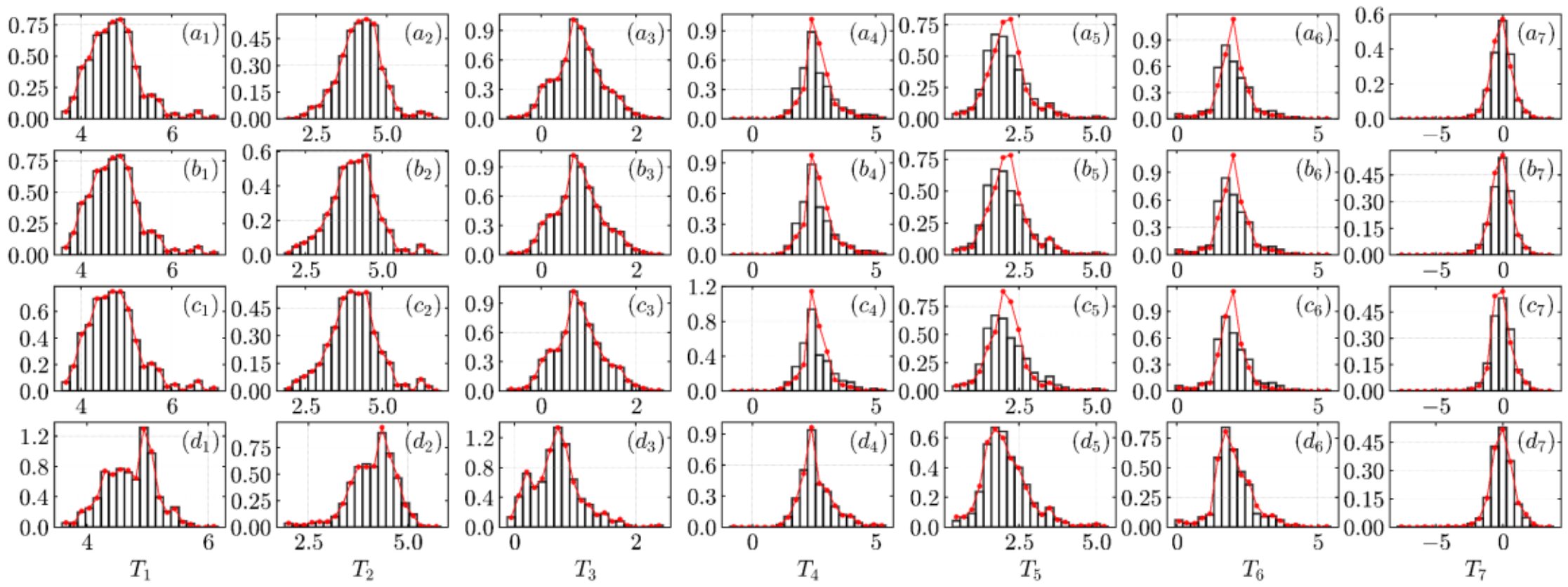

**Fig. 8.** Comparison of the MICE datasets (red dotted lines with markers) and the original datasets (black histograms). From top to bottom are the original dataset and Subsets 1 to 3.

As shown in the histograms of Figure 8, the overall shapes of the distributions remain consistent across the original dataset and Subsets 1 to 3. Minor differences can be observed in the distributions of variables $T_4$, $T_5$, and $T_6$ for the original dataset and Subsets 1 and 2 (Figures 8($a_4$)-($a_6$), 8($b_4$)-($b_6$), and 8($c_4$)-($c_6$)), where slight shifts in the density and a slight increase in peak height are noticeable. For other variables, the MICE and original data distributions are nearly indistinguishable.

Notably, as the data subsets progress from the full dataset to Subset 1 and then to Subset 3, the difference between the MICE and original data distributions decreases. This trend suggests that the MICE method becomes increasingly robust in imputing missing values as the number of available features increases. Across all subsets, the MICE data consistently resemble the original data, with only minimal distributional differences observed.

***3.3 Miss Forest (MF)***

MF is a variant of MICE and, like MICE, follows an iterative modelling approach, using known features to impute missing values. The key difference between MF and MICE lies in the imputation model: MICE uses traditional regression models, while MF replaces these with random forest models. As a non-parametric machine learning method, random forest is capable of capturing complex non-linear relationships between variables, making MF particularly effective for data with complicated interactions. Unlike MICE, which generates multiple imputed datasets, MF typically performs a single imputation. Both methods are widely used for missing data imputation. In this section, we will discuss the performance of the MF method when applied to impute the Clay/10/7490 database.

Like the analysis process in the previous section, we applied the MF method to impute missing values for each dataset, including the original database and Subsets 1-3. These datasets were used to investigate how different known data features impact the inference of missing data. Initially, missing values were imputed using the means of the variables to ensure the dataset was complete for model training. The algorithm then iteratively updates the missing values by training a random forest model on the available data. The process stops when the maximum change in imputed values between iterations falls below a predefined convergence threshold, indicating convergence. Key hyperparameters include the maximum number of iterations (set as 10 in this study), the convergence threshold ($10^{-4}$), the number of trees in the random forest (80), and the maximum depth of the trees (100).

Figure 9 presents a series of histograms comparing different variables ($T_1$ to $T_7$) across each subset. The black bars correspond to the original data, while the red bars represent the data imputed using the MF method. In Figures 9($a_4$), 9($b_4$), and 9($c_4$), the imputed data appear more concentrated than the original data distribution. For variables $T_4$, $T_5$, and $T_6$, slight differences in the histogram density are observed between the original database and Subsets 1 and 2. Notably, in Subset 3, the distribution of the imputed data closely matches that of the original data, consistent with the results obtained using the MICE method. Overall, in most cases, the distribution of imputed values aligns closely with the original data, indicating that the imputation process is generally successful.

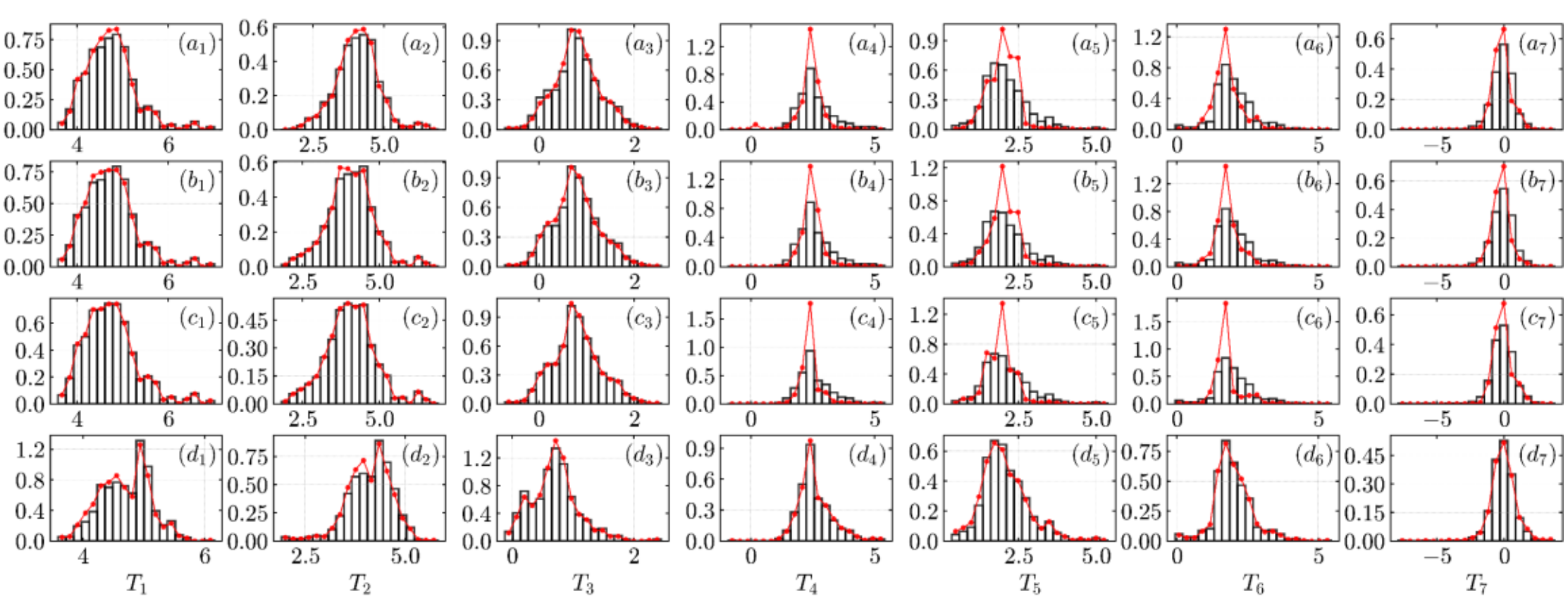

**Fig. 9.** Comparison of the MF datasets (red dotted lines with markers) and the original datasets (black histograms). From top to bottom are the original dataset and Subsets 1 to 3.

For better visualization and comparison, Figure 10 displays a series of violin plots showcasing the performance of the MN, MICE, and MF imputation methods on the Clay/10/7490 database. Each violin plot represents the probability distribution of the dataset after imputation, with the central black bar representing the quartiles of a corresponding box plot, while the white bar denotes the median.

In Figure 10, the distribution plots of variables $T_4$-$T_7$ clearly show that the database imputed by the MN database is more concentrated compared to the original database. Additionally, the database imputed by the MF method (hereafter referred to as the MF database) exhibits some extreme values and outliers, whereas the MICE method produces a more reliable database. This is primarily because, in the MICE method, the database is imputed 10 times, generating 10 different versions that are subsequently combined to form the final database.

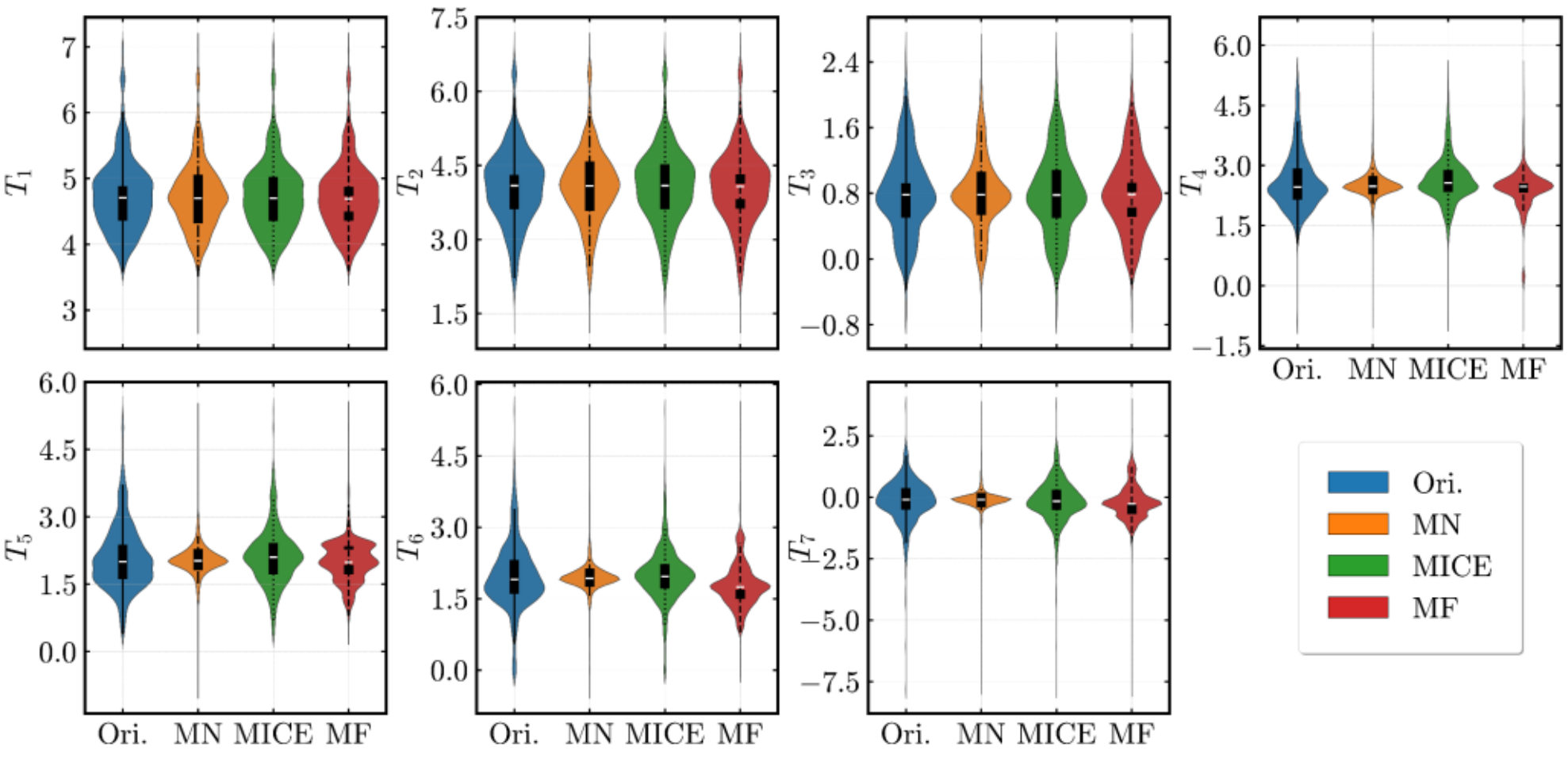

**Fig. 10.** Comparison of the distributions of original and imputed values (MN, MICE, MF) for each variable in the CLAY/10/7490 database

To quantitatively assess the similarity between the original and imputed data distributions, we performed the Kolmogorov-Smirnov (K-S) test, which evaluates the maximum difference between two cumulative distribution functions (CDFs). In the K-S test, smaller $D$ values and larger $p$-values (typically $p > 0.05$) indicate a closer match between distributions. As shown in Table 4, the MN method yields significantly higher $D$ values for $T_4$-$T_7$ (0.222-0.290, $p < 0.001$), indicating overly concentrated distributions. The MF method shows moderate deviation ($D$ = 0.166-0.211, $p < 0.05$). In contrast, MICE achieves the best performance with low $D$ values (e.g., 0.044 for $T_7$) and high $p$-values ($p > 0.05$), effectively preserving the original distribution characteristics. The poorer results for $T_4$-$T_6$ across all methods ($p < 0.0001$) are mainly due to their high missing rates (>85%), making accurate recovery difficult.

**Table 4.** K-S test results comparing original and imputed distributions

| $T_i$ | MN | MICE | MF |
|---|---|---|---|
| $T_1$ | $D$=0.040, $p$=0.002 | $D$= 0.009, $p$= 0.986 | $D$=0.029, $p$=0.048 |
| $T_2$ | $D$=0.033, $p$=0.011 | $D$= 0.005, $p$= 1.000 | $D$=0.023, $p$=0.157 |
| $T_3$ | $D$=0.064, $p$=0.000 | $D$= 0.005, $p$= 1.000 | $D$=0.030, $p$=0.036 |
| $T_4$ | $D$=0.222, $p$=0.000 | $D$= 0.136, $p$= 0.000 | $D$=0.200, $p$=0.000 |
| $T_5$ | $D$=0.256, $p$=0.000 | $D$= 0.097, $p$= 0.000 | $D$=0.185, $p$=0.000 |
| $T_6$ | $D$=0.289, $p$=0.000 | $D$= 0.117, $p$= 0.000 | $D$=0.211, $p$=0.000 |
| $T_7$ | $D$=0.274, $p$=0.000 | $D$= 0.044, $p$= 0.070 | $D$=0.166, $p$=0.000 |

Overall, as shown in Figure 10 and Table 4, the MICE database most closely aligns with the original database, followed by the MF database and then the MN database. However, these findings must be interpreted with caution, as statistical similarity represents only a necessary (not sufficient) criterion for evaluating imputation quality. Since the true values of missing data are unknown, it is impossible to determine which method is truly better. While preserving distributions and correlations offers useful insights, it does not guarantee that individual imputed values are accurate. This issue is especially critical for variables with high missingness (e.g., CPTU parameters with over 90% missing), where the imputed values reflect the predictions of the imputation method rather than the true data characteristics. Given these limitations, a more meaningful evaluation is the effect of each imputation method on target-variable prediction, which is examined in the following section.

## 4. Two Probabilistic Indirect Models

### *4.1 Probabilistic extreme gradient boosting (PXGB)*

Extreme gradient boosting (XGBoost) is a widely used and efficient ensemble learning algorithm built on a flexible gradient boosting framework, well-suited for structured data prediction tasks. A key advantage of XGBoost is its native ability to handle missing data during training. This eliminates the need for explicit imputation and simplifies the data preprocessing pipeline. Therefore, we can directly compare the performance of XGBoost applied to the original datasets versus the imputed datasets to evaluate the effectiveness of the imputation strategies (MN, MICE, and MF) discussed in Sections 2 and 3.

In this study, we extend XGBoost into a probabilistic framework, referred to as probabilistic XGBoost (PXGB), by assuming that the target variable follows a normal distribution. To enable uncertainty quantification in predictions, we adopt the negative log-likelihood (NLL) as the loss function for model optimization. This process can be formally expressed as:

$$T_8^{\text{pre}} \sim \mathrm{N}\left(\mu^{\text{pre}}, \left(\sigma^{\text{pre}}\right)^2\right) \tag{12}$$

$$\left(\mu^{\text{pre}}, \sigma^{\text{pre}}\right) = f_{\text{PXGB}}(T_1, T_2, \ ..., T_7) \tag{13}$$

$$L = -\sum_{i=1}^{N} \log\left(p(y_i |\ x_i; \boldsymbol{\theta})\right) \tag{14}$$

where $T_8^{\text{pre}}$ represents the predicted probabilistic distribution for the transformed target variable $s_{\text{u}}(\text{mob})/\sigma'_{\text{v}}$, $L$ is the loss function (NLL), $N$ is the number of data points, $p(y_i|x_i;\boldsymbol{\theta})$ is the probability of the target variable $y_i$ given the input $x_i$ and model parameters $\boldsymbol{\theta}$.

### *4.2 Multi-head attention-based probabilistic neural network (MHA-PNN)*

Attention mechanism is inspired by how humans focus on important details in a scene while ignoring irrelevant information. Multi-head attention (MHA) extends this concept by projecting queries ($\boldsymbol{Q}$), keys ($\boldsymbol{K}$), and values ($\boldsymbol{V}$) into multiple low-dimensional subspaces. This allows for parallelized feature learning and enhances the extraction of information from multiple perspectives. Therefore, in this study, the MHA layer is incorporated into a traditional neural network model, with the probabilistic distributions of the target as the output. This model is referred to as the multi-head attention-based probabilistic neural network (MHA-PNN). The architecture of the MHA-PNN model comprises four key modules, including an input layer, a hidden layer, an MHA layer, and an output layer. The process begins with the input layer, which receives a series of transformed input features [$T_1$, $T_2$, ..., $T_7$]. These features are subsequently mapped into the hidden layer, where they are represented as corresponding vectors [$d_1$, $d_2$, ..., $d_d$]. Next, the hidden representations are processed through the MHA layer, yielding output vectors [$o_1$, $o_2$, ..., $o_j$]. These outputs are then normalized to obtain the final representations [$d_1$, $d_2$, ..., $d_j$], which serve as the predictors for estimating the probability distribution of the target variable. Additionally, to mitigate overfitting, dropout regularization is incorporated into both the hidden layer and the MHA layer. The overall architecture of the MHA-PNN model is illustrated in Figure 11.

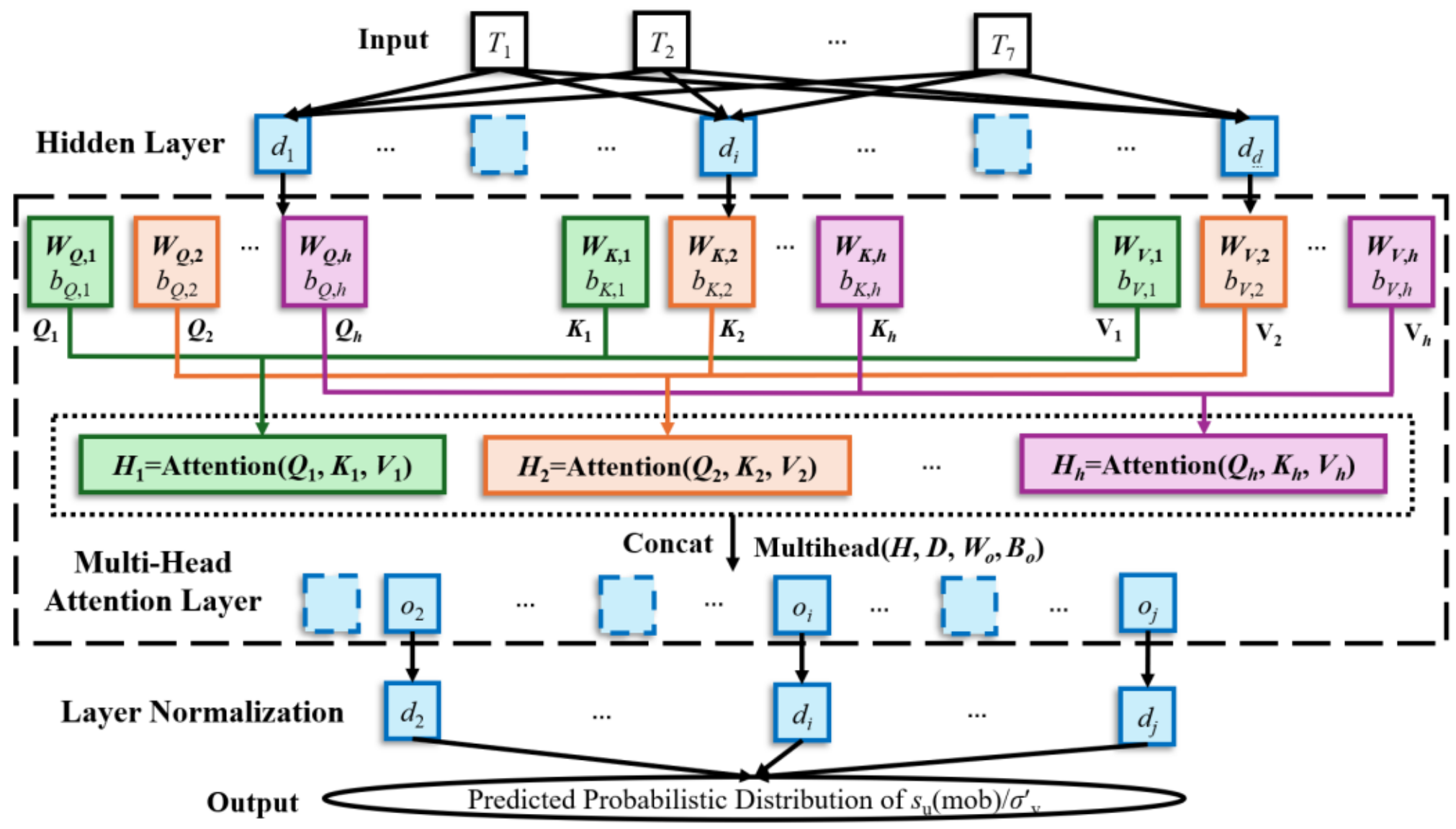

**Fig. 11.** The architecture of the MHA-PNN model

As shown in Figure 11, the key aspect of the model lies in the implementation of the MHA layer and the subsequent probabilistic distribution prediction. For the MHA layer, linear transformations are applied to generate the $\boldsymbol{Q}$, $\boldsymbol{K}$, and $\boldsymbol{V}$ matrices. Specifically, each attention head independently projects the hidden representations using different trainable weight matrices, ensuring diverse feature extraction and capturing distinct aspects of the input information. The transformation for each attention head is given by the following equations:

$$\begin{aligned} \boldsymbol{Q}_h &= \boldsymbol{DW}_{Q,h} + b_{Q,h} = \left[q_{h,1}, q_{h,2}, \cdots, q_{h,d}\right] \\ \boldsymbol{K}_h &= \boldsymbol{DW}_{K,h} + b_{K,h} = \left[k_{h,1}, k_{h,2}, \cdots, k_{h,d}\right] \\ \boldsymbol{V}_h &= \boldsymbol{DW}_{V,h} + b_{V,h} = \left[v_{h,1}, v_{h,2}, \cdots, v_{h,d}\right] \end{aligned} \tag{15}$$

where $\boldsymbol{W}_{Q,h}$, $\boldsymbol{W}_{K,h}$, and $\boldsymbol{W}_{V,h}$ represent the weight matrices for the $\boldsymbol{Q}_h$, $\boldsymbol{K}_h$, and $\boldsymbol{V}_h$, respectively, while $b_{Q,h}$, $b_{K,h}$, and $b_{V,h}$ are the corresponding bias terms. Here, $h$ refers to the number of attention heads, $d$ is the number of neurons in the hidden layer, and $\boldsymbol{D}$ represents the hidden representations. Each attention head functions independently, capturing distinct aspects of the input information. It leverages scaled dot-product attention to compute the relevance between features, with the specific computation given as follows:

$$\boldsymbol{H}_h = \text{Attention}(\boldsymbol{Q}_h, \boldsymbol{K}_h, \boldsymbol{V}_h) = \text{softmax}\left(\frac{\boldsymbol{Q}_h \cdot \boldsymbol{K}_h^{\ T}}{\sqrt{d_k}}\right) \cdot \boldsymbol{V}_h \tag{16}$$

where $d_{\mathrm{k}}$ is the dimension of $\boldsymbol{K}_h$, which is used to scale the dot product between $\boldsymbol{Q}_h$ and $\boldsymbol{K}_h$, the softmax function is the normalized exponential function, and $\boldsymbol{H}_h$ represents the output of the attention mechanism for the $h$-th attention head. The outputs from these heads are then concatenated (Concat) and linearly transformed to yield the final multi-head attention output. To preserve the original features, we incorporate a residual connection in the multi-head attention mechanism. The final formula is as follows:

$$\boldsymbol{O} = \text{Multihead}(\boldsymbol{H}, \boldsymbol{W}_o, \boldsymbol{B}_o, \boldsymbol{T}) = \text{concat}\left(\boldsymbol{H}_1, \boldsymbol{H}_2, \cdots, \boldsymbol{H}_h\right)\boldsymbol{W}_o + \boldsymbol{B}_o + \boldsymbol{T} \tag{17}$$

where $\boldsymbol{W_o}$ is the weight matrix used to combine the information from different attention heads, and $\boldsymbol{B_o}$ is the corresponding bias term. The outputs of the MHA layer are normalized and then passed to the output layer, where a probabilistic modelling approach is used to obtain the probability distribution of the target variable. In the output layer, as shown in Eqs (12)-(14) and consistent with the PXGB model, the output is assumed to follow a normal distribution, and the NLL is adopted as the loss function. Notably, the softplus function is used to ensure that the predicted standard deviation remains positive, with a small constant ($10^{-6}$) added for numerical stability.

### *4.3 Hyperparameter configuration*

Hyperparameter optimization is essential for improving the performance of machine learning models and ensuring accurate predictions of the target variable. To minimize the influence of hyperparameter settings on the performance of the PXGB and MHA-PNN models, we employ Optuna, a hyperparameter optimization technique, to automate the identification of the optimal hyperparameters. Specifically, for the MHA-PNN model, we focus on optimizing the following hyperparameters: the number of attention heads ($N_{att}$), the key dimension ($d_k$), the number of units in the dense layer ($N_{units}$), the dropout rate ($r_d$) in both the dense and MHA layers, the learning rate ($\eta$), and the activation function (*act*). The initial search ranges for these hyperparameters are as follows: $N_{att}$ and $d_k$ from 2 to 20, $N_{units}$ from 20 to 200, $r_d$ from 0.2 to 0.5, $\eta$ from $10^{-4}$ to $10^{-1}$, and *act* options being ReLU, Sigmoid, and Tanh. For the PXGB model, the optimized hyperparameters include the learning rate ($\eta$), maximum tree depth ($D_{max}$), minimum child weight ($w_{min}$), regularization term ($\gamma$), feature subsampling ratio ($r_{fea}$), and instance subsample ratio ($r_{ins}$). Their corresponding search spaces are: $\eta$ from $10^{-4}$ to $10^{-1}$, $D_{max}$ from 1 to 20, $w_{min}$ from 1 to 20, $\gamma$ from 0 to 0.2, $r_{fea}$ from 0.5 to 1, and $r_{ins}$ from 0.8 to 1. For each dataset, a total of 10 trials were performed to identify the best-performing configurations. The optimal hyperparameter configurations are summarized in Table 5.

**Table 5.** Optimal hyperparameter configurations for the PXGB and MHA-PNN models

| $T_i$ | PXGB | | | | MHA-PNN | | |
|---|---|---|---|---|---|---|---|
| | Ori. | MN | MICE | MF | MN | MICE | MF |
| $N_{att}$ | - | - | - | - | 20 | 7 | 9 |
| $d_k$ | - | - | - | - | 19 | 2 | 13 |
| $N_{units}$ | - | - | - | - | 184 | 84 | 152 |
| $r_d$ | - | - | - | - | 0.31 | 0.23 | 0.41 |
| *act* | - | - | - | - | ReLU | Tanh | ReLU |
| $\eta$ ($\times 10^{-2}$) | 4.29 | 2.86 | 6.87 | 8.15 | 0.012 | 0.21 | 0.10 |
| $D_{max}$ | 17 | 19 | 11 | 20 | - | - | - |
| $w_{min}$ | 10 | 10 | 8 | 14 | - | - | - |
| $\gamma$ | 0.16 | 0.14 | 0.16 | 0.18 | | | |
| $r_{fea}$ | 0.97 | 0.78 | 0.79 | 0.87 | - | - | - |
| $r_{ins}$ | 0.86 | 0.86 | 0.88 | 0.86 | - | - | - |

The results show that the optimal hyperparameters vary across datasets, emphasizing the necessity of dataset-specific tuning. Leveraging these optimal models enhances the reliability and credibility of subsequent comparative analyses, ensuring that the findings are not affected by hyperparameter configurations.

### *4.4 Performance comparison and uncertainty quantification*

Based on the optimal hyperparameters determined in the previous section, we constructed the corresponding optimal model for each dataset. To ensure reliable evaluation, we adopted stratified 5-fold cross-validation, which maintains consistent label distributions across all folds. This approach prevents the absence of certain label types in any given fold, ensuring that model performance is assessed fairly and comprehensively. The random seed was fixed at 42 during data splitting and training to ensure reproducibility. In each fold, the model was trained on four folds and validated on the remaining one. Figures 12-13 present the validation results for each fold, comparing the PXGB and MHA-PNN models across different datasets.

Figure 12 presents the prediction results of the PXGB model under different labels, ranging from 1 to 7 for different rows. From left to right, each column displays the validation results for folds 1 through 5. In this figure, green, blue, red, and purple represent the predictions of models trained on the original dataset, the MN dataset, the MICE dataset, and the MF dataset, respectively. It is evident that the predictions across the five folds are visually consistent, demonstrating the model's robustness and stability across different data splits. Moreover, the predictions based on the imputed datasets (MN, MICE, and MF) generally exhibit improved accuracy, as the results using the original dataset (in green) deviate more noticeably from the diagonal line. This visual deviation indicates larger prediction errors, thereby validating the effectiveness of the data imputation methods in enhancing model performance.

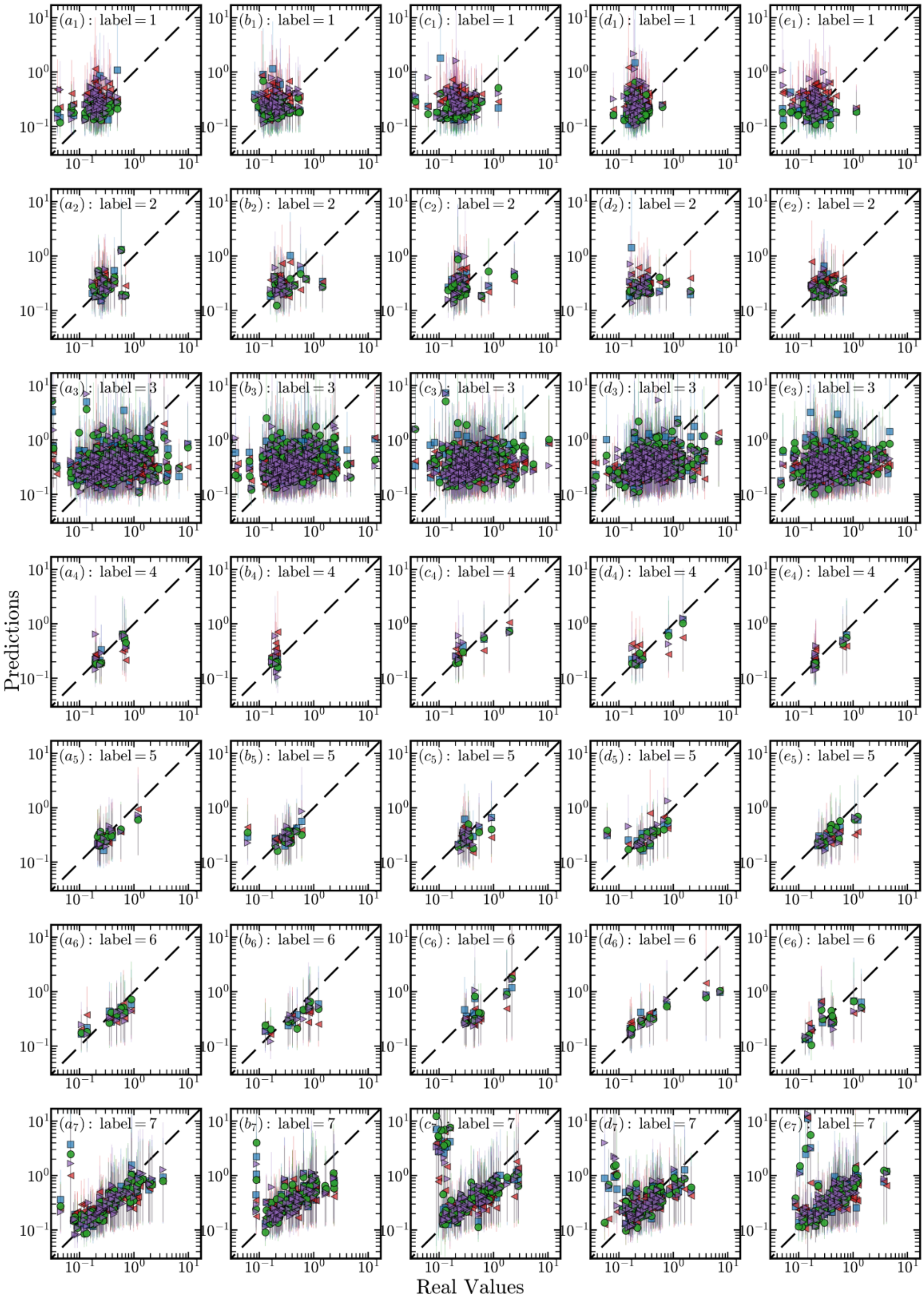

**Fig. 12.** PXGB model predictions across five folds using original and imputed datasets (left to right: folds 1-5; green = original dataset, blue = MN dataset, red = MICE dataset, purple = MF dataset).

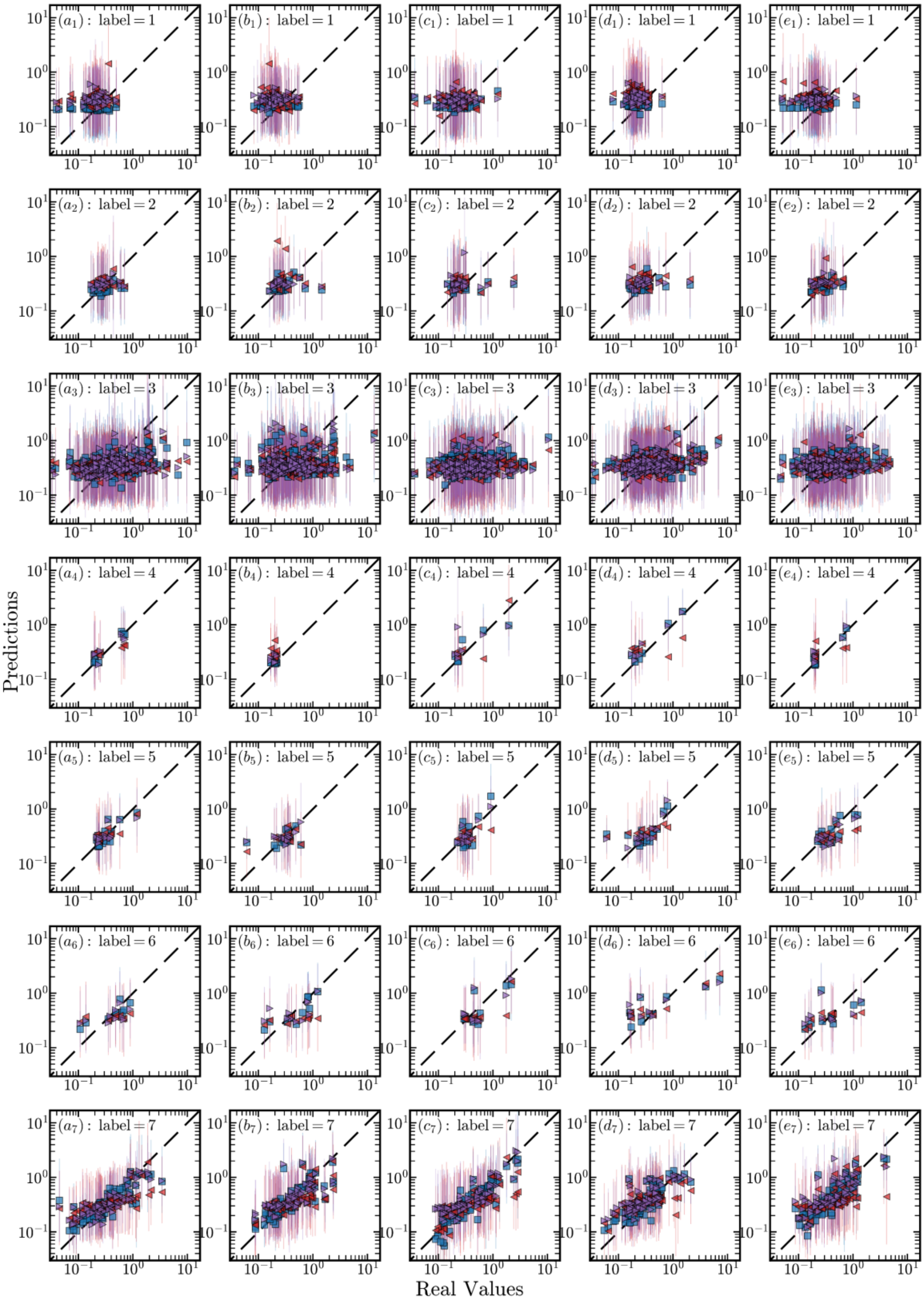

**Fig. 13.** MHA-PNN model predictions across five folds using imputed datasets (left to right: folds 1-5; blue = MN dataset, red = MICE dataset, purple = MF dataset).

Figure 13 presents the prediction performance of the MHA-PNN model under five-fold cross-validation, using only the imputed datasets (blue for MN, red for MICE, and purple for MF). Similar to the PXGB model, the prediction results appear visually consistent across all five folds, demonstrating the stability of both proposed models. Notably, compared to PXGB, the MHA-PNN model shows fewer points deviating from the diagonal line, indicating improved prediction accuracy. However, as illustrated in Figures 12 and 13, as well as in the earlier MN-based prediction results shown in Figure 4, when the number of available input features is less than or equal to three, the model exhibits consistently poor predictive performance, regardless of whether the original or imputed dataset is used. To ensure a fair comparison between models, these low-information input cases (i.e., with ≤3 input features) are excluded from the subsequent evaluations, and only results with more than three input features are considered.

In this study, three commonly used evaluation metrics are employed to assess prediction accuracy: root mean square error (*RMSE*), coefficient of determination ($R^2$), and mean absolute percentage error (*MAPE*). Additionally, to further evaluate the quality of uncertainty quantification, two additional metrics are employed: the width of the conditional interval ($w_{\mathrm{CI}}$) and the coverage rate (*CR*), which are defined as follows:

$$CR = \frac{1}{N}\sum_{i=1}^{N}\mathbf{1}\left(y_i^{\mathrm{real}} \in \left[y_{i,\mathrm{L}}^{\mathrm{pre}}, y_{i,\mathrm{U}}^{\mathrm{pre}}\right]\right) \tag{18}$$

$$w_{\mathrm{CI}} = \frac{1}{N}\sum_{i=1}^{N}\left[y_{i,\mathrm{U}}^{\mathrm{pre}} - y_{i,\mathrm{L}}^{\mathrm{pre}}\right] \tag{19}$$

where $N$ is the total number of dataset samples, $y_i^{\mathrm{real}}$ denotes the real value of the $i$-th sample, $y_{i,\mathrm{M}}^{\mathrm{pre}}$, $y_{i,\mathrm{U}}^{\mathrm{pre}}$ and $y_{i,\mathrm{L}}^{\mathrm{pre}}$ represent the predicted median, upper, and lower bounds of the 95% confidence interval, respectively, and $\mathbf{1}(\cdot)$ is the indicator function, which equals 1 if the real value falls within the predicted 95% confidence interval, and 0 otherwise.

Among these metrics, $CR$ shows how often the true values fall inside the predicted range. A higher $CR$ means better coverage. $w_{CI}$ indicates the average width of the predicted range. A smaller $w_{CI}$ means the model is more confident, as long as $CR$ remains high enough.

**Table 6.** Evaluation metrics of PXGB, MHA-PNN, and MN-based prediction (MN-pred.) models on the original and imputed datasets

| Model (Dataset) | *MAPE* | *RMSE* | $R^2$ | $w_{CI}$ | *CR* |
|---|---|---|---|---|---|
| MN-pred. (Ori.) | 0.48 | 0.59 | -0.06 | 0.72 | 0.80 |
| PXGB (Ori.) | 1.72 | 1.17 | -5.62 | 434.62 | 0.97 |
| PXGB (MN) | 1.22 | 0.78 | -1.31 | 24.86 | 0.97 |
| PXGB (MICE) | 1.00 | 0.76 | -1.29 | 28.81 | 0.93 |
| PXGB (MF) | 1.19 | 0.78 | -1.40 | 44.95 | 0.95 |
| MHA-PNN (MN) | 0.34 | 0.37 | 0.55 | 1.03 | 0.94 |
| MHA-PNN (MICE) | 0.46 | 0.45 | 0.32 | 1.58 | 0.98 |
| MHA-PNN (MF) | 0.46 | 0.39 | 0.50 | 1.47 | 0.97 |

Table 6 provides a comprehensive comparison of the PXGB, MHA-PNN, and MN-based prediction models across both original and imputed datasets. Among the evaluated metrics, the PXGB model using the original dataset shows the poorest performance, with the least favourable metrics ($MAPE$ = 1.72, $RMSE$ = 1.17, $R^2$ = -5.62, $w_{CI}$ = 434.62), reflecting its limited predictive capability when handling highly incomplete data. In contrast, notable improvements are observed for PXGB when applied to imputed datasets, particularly the MICE dataset, where the performance improves significantly ($MAPE$ = 1.00, $RMSE$ = 0.76, $R^2$ = -1.29, $w_{CI}$ = 28.81), demonstrating the positive impact of effective data imputation. Specifically, compared to the PXGB model on the original dataset, the PXGB model on the MICE dataset reduced $MAPE$ by approximately 42%, $RMSE$ by 35%, increased $R^2$ from -5.62 to -1.29, and decreased $w_{CI}$ by about 93%, while $CR$ remained stable.

Additionally, the best overall results are achieved by the MHA-PNN model trained on the MN dataset, with the lowest *MAPE* (0.34) and *RMSE* (0.37), as well as the highest $R^2$ (0.55) and strong uncertainty metrics ($w_{CI}$ = 1.03, *CR* = 0.94). On the MN dataset, the MHA-PNN model substantially outperformed the PXGB model, reducing *MAPE* by approximately 72%, *RMSE* by 53%, increasing $R^2$ from -1.31 to 0.55, and decreasing $w_{CI}$ by about 96%, while *CR* remained stable. The results collectively underscore that predictive performance is highly sensitive to both the choice of prediction model and the imputation method. Advanced architectures such as MHA-PNN, when paired with appropriate imputation strategies, can substantially enhance the accuracy and reliability of predictions, particularly in data-scarce geotechnical settings.

To maintain consistency, the three metrics (*MAPE*, *RMSE*, and $w_{CI}$) were transformed by taking their reciprocals, ensuring that higher values uniformly indicate better performance. Subsequently, all five metrics were normalized using min-max scaling. This allows for a more intuitive comparison, as shown in Figure 14, which presents a radar chart of the normalized evaluation metrics across various models and datasets. In this figure, the blue colour scheme, ranging from dark to light, corresponds to the PXGB model results on the original (Ori.), MN, MICE, and MF datasets, respectively; the purple colour scheme, also from dark to light, represents the MHA-PNN model applied to the MN, MICE, and MF datasets; the green dotted line with markers denotes the MN-based prediction (MN-pred.) model on the original dataset. It is evident that the dark blue line with markers (PXGB on the original dataset) exhibits the poorest overall performance, indicated by the smallest radar area. Conversely, the purple dotted lines with markers representing MHA-PNN consistently achieve superior performance, occupying the outermost regions of the chart.

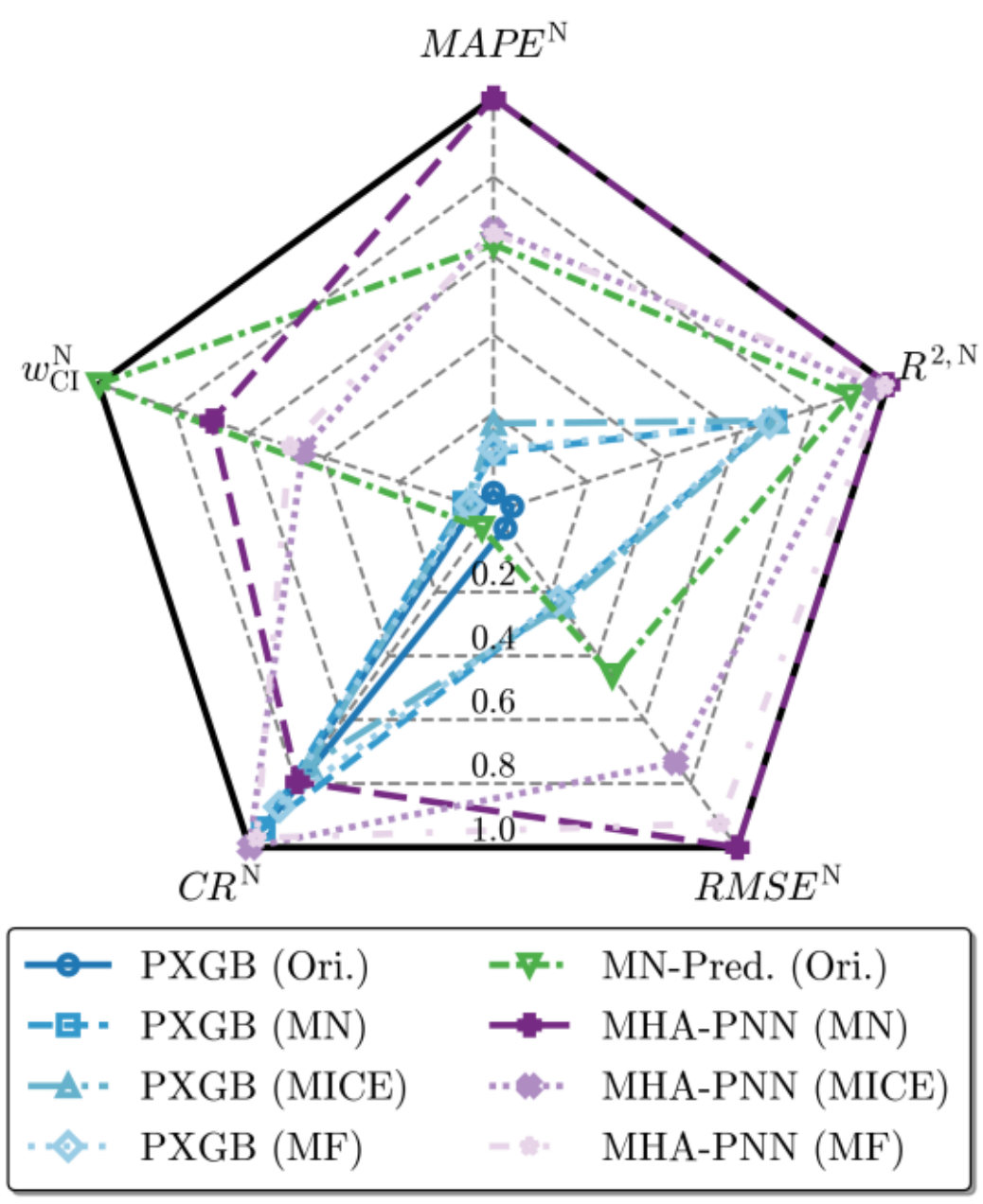

**Fig. 14.** Radar chart of normalized evaluation metrics for different models and datasets

Overall, these findings confirm that MHA-PNN significantly outperforms both PXGB and MN-pred. models in predictive accuracy and uncertainty quantification. Among the imputation methods applied to the MHA-PNN model, the MN approach offers the best trade-off between prediction accuracy and reliable uncertainty quantification.

## 5. Conclusions

The prediction of $s_u$ is essential for applications such as slope stability analysis and foundation bearing capacity evaluation. However, traditional estimation methods, including laboratory tests and in-situ techniques, are often subject to significant uncertainty. Moreover, the sparsity and variability of geotechnical data greatly limit the development of robust predictive models. To address these issues, this study proposes a probabilistic modelling framework that combines statistical and machine learning techniques to generate more reliable predictions of $s_u$ along with uncertainty quantification. The framework is developed using the CLAY/10/7490 database, which presents substantial challenges due to its low completion rate (around 34.1%) and high variability. To mitigate data incompleteness, we systematically explore the effects of different imputation strategies (MN, MICE, and MF) and develop an MN-based prediction model, a probabilistic XGBoost (PXGB) model, and a neural network variant for predicting probabilistic distributions of $s_u$.

Initially, the MN-based prediction model was applied to the CLAY/10/7490 database to predict the probabilistic distribution of $s_u(\text{mob})/\sigma'_v$ using seven input features, including Atterberg limit parameters and CPTU parameters. The results revealed that the number of input features had a significant impact on prediction accuracy. Particularly, when fewer than four features were used, the model predictions became largely unreliable. Subsequently, the MN, MICE, and MF imputation methods were applied to address the data incompleteness. Statistical analyses revealed that although the imputed results remained sensitive to the number of input features, the distributions of the completed databases closely matched those of the original database. However, we emphasize that imputed values should not be interpreted as accurate reconstructions of the true, unobserved data; therefore, site-specific investigations remain essential for reliable decision-making.

The effectiveness of the MN, MICE, and MF imputation methods was validated by comparing the performance of the PXGB model trained on imputed datasets against its performance on the original incomplete dataset. To enhance the model's capability in extracting information from limited inputs, an MHA mechanism was incorporated into the traditional ANN architecture. By using the predicted probability distribution of the target variable as the output, a novel MHA-PNN model was developed. Comprehensive evaluations using $RMSE$, $R^2$, $MAPE$, $CR$, and $w_{CI}$ demonstrate that the proposed MHA-PNN consistently achieves lower prediction errors, higher coverage rates, and narrower uncertainty intervals. For example, on the MN dataset, the MHA-PNN model outperformed PXGB, reducing *MAPE* by approximately 72%, *RMSE* by 53%, increasing $R^2$ from -1.31 to 0.55, and decreasing $w_{\mathrm{CI}}$ by about 96%, while *CR* remained stable.

In summary, the MHA-PNN model exhibits clear advantages over the MN-based prediction and PXGB models in both predictive accuracy and uncertainty quantification across various datasets. This model effectively captures complex feature interactions, improves prediction accuracy under sparse input conditions, and provides well-calibrated uncertainty estimates. However, we emphasize that this model is not a substitute for site-specific predictions. We explicitly recommend that engineers treat the model outputs as preliminary guidance and supplement them with local testing whenever feasible. In particular, when applying the model to geologically distinct or sensitive conditions (e.g., structured or soft clays), incorporating site-specific data is mandatory for reliable analysis.

**Acknowledgements**

The authors would like to thank the members of the TC304 Committee on Engineering Practice of Risk Assessment & Management of the International Society of Soil Mechanics and Geotechnical Engineering for developing the database 304dB used in this study and making it available for scientific inquiry. We also wish to thank Dr. Jianye Ching for contributing this CLAY/10/7490 database to the TC304 compendium of databases.